\documentclass{article}

\usepackage{PRIMEarxiv}
\usepackage{amsmath}
\usepackage[ruled,vlined]{algorithm2e}
\usepackage{forest}
\usepackage{tabularx,booktabs}
\newcolumntype{Y}{>{\centering\arraybackslash}X}
\usepackage{graphicx}
\usepackage{subcaption}
\usepackage[utf8]{inputenc} % allow utf-8 input
\usepackage[T1]{fontenc}    % use 8-bit T1 fonts
\usepackage{hyperref}       % hyperlinks
\usepackage{url}            % simple URL typesetting
\usepackage{booktabs}       % professional-quality tables
\usepackage{amsfonts}       % blackboard math symbols
\usepackage{nicefrac}       % compact symbols for 1/2, etc.
\usepackage{microtype}      % microtypography
\usepackage{lipsum}
\usepackage{fancyhdr}       % header
\usepackage{graphicx}       % graphics
\graphicspath{{media/}}     % organize your images and other figures under media/ folder

\title{Why Are You Weird?\\Infusing Interpretability in Isolation Forest for Anomaly Detection}
\author{
  Nirmal Sobha Kartha \textsuperscript{\rm 1}, Clement Gautrais \textsuperscript{\rm 2} and Vincent Vercruyssen \textsuperscript{\rm 2} \\
  KU Leuven \\
  \textsuperscript{\rm 1} \texttt{nirmal.nsk.kartha@gmail.com}, \textsuperscript{\rm 2} \texttt{firstname.lastname@kuleuven.be} \\
}

\begin{document}
\maketitle

\begin{abstract}
%%% Context: (Why now? Give background)
Anomaly detection is concerned with identifying examples in a dataset that do not conform to the expected behaviour. 
%%% Need (Why you: Why is this useful to the reader?)
While a vast amount of anomaly detection algorithms exist, little attention has been paid to explaining \emph{why} these algorithms flag certain examples as anomalies. However, such an explanation could be extremely useful to anyone interpreting the algorithms' output.
%%% Task (Why me/us: What are we solving?)
This paper develops a method to explain the anomaly predictions of the state-of-the-art Isolation Forest anomaly detection algorithm.
%%% Object (Why this document: How are we going to solve this?)
The method outputs an \emph{explanation vector} that captures how important each attribute of an example is to identifying it as anomalous.
%%% Findings (What? We found that ...)
A thorough experimental evaluation on both synthetic and real-world datasets shows that our method is more accurate and more efficient than most contemporary state-of-the-art explainability methods.
%%% Conclusions (So what? What do these results imply?)
%%% Perspective: (What now? What does the future look like?)
\end{abstract}

% keywords can be removed

\section{Introduction}

Anomaly detection is a crucial data mining task as it deals with identifying examples in a dataset that do not conform to the expected behaviour (also called anomalies). Detecting anomalies is essential in mission-critical domains such as network intrusion detection~\cite{Garcia-Teodoro2009-sj}, credit scoring and fraud detection~\cite{Thiprungsri2011-jz,Bolton2001-fl}, or patient screening and monitoring~\cite{Lin2005-xz}.

Anomaly detection algorithms usually work in a completely \emph{unsupervised} manner~\cite{campos2016evaluation}. They do not make use of label information to identify anomalous examples in a dataset. Such algorithms work by assigning each example an anomaly score derived from its attributes. High scores indicate anomalousness while low scores indicate normality. By thresholding these scores, we obtain a binary label (anomaly or normal).

While a variety of algorithms exist to compute different types of anomaly scores, surprisingly little attention has been devoted to developing methods that can explain \emph{why} a particular score is assigned to an example. However, such an \emph{anomaly explanation} could be instrumental for domain experts interpreting the algorithms' output. For instance, an explanation of which factors led to flagging a network activity as anomalous, i.e., an intrusion, can help the expert design a better security system.
Informally, model explainability refers to the degree to which the human can understand the cause of a model decision~\cite{Miller2019-qb}.

The ability to explain the output of an anomaly detection algorithm will also engender trust in its predictions~\cite{molnar2019}. In this paper, we tackle the challenge of explaining the output of the \emph{Isolation Forest} (\textsc{iForest}) anomaly detection algorithm~\cite{Liu2008-bj}. Although this algorithm often achieves state-of-the-art performance~\cite{domingues2018comparative}, it is a black-box model with little to no model interpretability.
Concretely, we make the following contributions:
\begin{itemize}
    \item We contribute an algorithm that exploits the structure of the \textsc{iForest} method to explain its predictions.
    \item We conduct an experimental study to evaluate the quality and speed of the generated explanations and show that our method performs on par with the state-of-the-art while being an order of magnitude faster.
\end{itemize}

\section{Preliminaries}

We briefly describe how the \textsc{iForest} algorithm works~\cite{Liu2008-bj}. This algorithm's main assumption is that anomalies are easier to \emph{isolate} from the rest of the data than normals whereby the isolation is achieved by randomly and recursively splitting the data.
The \textsc{iForest} algorithm has two steps.
Given a dataset $X = \{x_1, \ldots, x_n\}$ with each example $x \in \mathbf{R}^d$, it first builds an ensemble of $T$ isolation trees. Each tree recursively splits a random subsample of size $\psi$ of the data using axis-parallel splits until each example ends up in its own leaf node or the tree depth limit is reached. In each split node, the split attribute and split value for that attribute are chosen randomly. 
Then, it computes an anomaly score $a(x)$ for each example $x$ based on the path lengths of the example traversing each tree and the size of the nodes it ends up in:
% add scoring mechanism?
\begin{align*}
    a(x) &= \frac{1}{T} \sum_{\text{trees}} (x \text{ leaf node depth} + \nu(x \text{ leaf node size})) \\
    \nu(i) &= 2 H(i) - 2
\end{align*}
with $H(i)$ the $i^{th}$ harmonic number.
The score can be thresholded to derive a binary label for each example. %(i.e., an example is normal or anomalous).

\section{Methodology}
\label{sec:method}

The problem we are trying to solve in this paper, is:
\begin{description}
    \item[Given:] A dataset $X = \{x_1, \ldots, x_n\}$ and an ensemble of $T$ Isolation trees built using this dataset.
    \item[Do:] Generate an explanation vector $w \in \mathbf{R}^d$ for each example $x$ that quantifies how important each attribute is to the predicted label of $x$.
\end{description}

Our explainability method is uniquely designed for the \textsc{iForest} anomaly detection algorithm.
The key insight behind our approach is that the importance of an attribute to predicting an example as anomalous, correlates with how instrumental that attribute is towards isolating the example. 
In an isolation tree, an attribute's power to isolate an example is captured by its ability to shorten the expected depth of the branch the example finally lies in. Intuitively, those attributes contribute more to the explanation of an example's predicted label.
The next sections present what shortening the expected branch depth means in greater detail.
%how many times that attribute is used to split the data within an isolation tree.
%

For a given example, our method outputs a vector $w \in \mathbf{R}^d$ capturing the importance of each attribute in predicting the example's label. The value in $w$ for each attribute represents the average path length shortening obtained by using this attribute. 
This average is computed over all the trees in the \textsc{iForest}.
A value close to 0 means that the attribute had little to no influence in making this data point anomalous. A negative value means that this attribute made the point look like a normal point. A positive value means that the attribute had an influence in making the point anomalous.
% \textcolor{red}{[say something more about how the method works high-level]}

\subsection{Intuitions}
\label{subsec:intuitions}
Our method leverages three intuitions about the structure of the \textsc{iForest} algorithm to explain its predictions.
First, the split attributes along an example's path in an isolation tree are instrumental in isolating it from the rest of the data. Thus, they are likely to explain the example's predicted label.
Second, anomalous examples are more likely to be isolated closer to the root, which is also the core premise of the \textsc{iForest} algorithm. This tells us to assign more weight to split attributes involved with larger parent node sizes,  when computing the explanation vector.
Finally, the most informative split nodes help shorten an example's path length in an isolation tree. This entails that split attributes and split values are more informative if they split the data in that node in an unbalanced manner and if they allocate the example we are explaining to the smaller child node.
Our method employs a weighting scheme that takes into account these three intuitions.

% Each tree in the \textsc{iForest} attempts to isolate examples from the rest of the data using random axis-parallel splits.
% - intuition of the tree weighting.
% - intuition about the node weighting

\subsection{Algorithm}
\label{subsec:ourapproach}

% Building on these observations, we propose the following methodology to make the IF produce local explanations for anomalous points of interest. Let us consider an already trained IF ensemble of Isolation trees, $IF = \{t_1, t_2, ..., t_T\}$, fit on the input dataset $D = \{x_1, x_2, ..., x_n\}$ where each sample, $x_i$ is represented by a \textit{d}-dimensional feature vector.
%, i.e., 
%x^{(i)}=[x^{(i)}_1, x^{(i)}_2, ..., x^{(i)}_d]^T$
% where $x^{(i)}_k \in \Re \; \forall i, k$ %are the values of $x^{(i)}$ along features/attributes $f_1, f_2, ..., f_d: d \in \Re$. \vincent{blue}{this is unnecessarily complex. We need to clean this up.} Each Isolation tree $t \in IF$ is a collection of nodes $\nu$ s.t. $t = \{\nu_1, \nu_2, ..., \nu_N\}$. The total number of trees in the forest, $T$, is a hyperparameter set to a default value of 100. Each Isolation tree, $t$, is assigned bootstrapped samples, $D_{t} \in D$ ($|D_{t}|=256$ by default). The maximum depth, $h_{lim}$, can either be manually inputted by the user or automatically set for each tree $t$ to $log_2(|D_{t}|) \; \forall t \in IF$.

Let $x \in \mathbf{R}^d$ be an example for which we want to explain its anomaly score output by the \textsc{iForest}. The \textsc{iForest}$\ = \{t_1, t_2, ..., t_T\}$ consists of an ensemble of $T$ isolation trees trained on the full dataset.
Algorithm \ref{alg:feature_importance} presents the computation of the importance of each attribute to the anomaly score. Its output is the explanation vector $w$.
\textbf{Line 2} iterates over all trees in the Isolation Forest. 
\textbf{Line 3} initiates the traversal of the isolation tree by starting at the root. 
\textbf{Lines 6 to 11} compute to which child node the example $x$ should go to. It does so by following the data split of the current node. 
\textbf{Line 12} computes the score of the considered split. It depends on both the parent node size and the current node size. Section~\ref{subsec:weights} explains this step in more detail.
\textbf{Line 13} attributes the score of the current node to the split attribute that is used in this node and adds it to the attribute importance vector $w$.
%\clement{red}{This is a ratio, why do we do an arithmetic mean and not a geometric one???}
After traversing all trees, the resulting vector $w$ is returned. This vector is in fact the explanation vector we set out to compute.

% each instance, $x^{(i)} \in D$, which are thresholded by a threshold value, either manually provided by the user or determined by a user-inputted contamination factor. Let $\hat{y^{(i)}}$ and $\hat{y^{(i)}_t}$ be the thresholded output predictions for each instance, $x^{(i)} \in D$, by $IF$ and each $t\in IF$ respectively. 

%\vincent{blue}{This is very complex and difficult to read. This needs a rewrite + maybe structured algorithm. Part of the issue is that the explanation mingles both the algorithmic components + how you would effectively implement it. So its neither fish nor fowl.}

\begin{algorithm}[h]
\SetAlgoLined
\LinesNumbered
\KwIn{An \textsc{iForest}=$\{t_1, t_2, ..., t_T\}$ and an example $x=[x_1, x_2, \ldots, x_d]$}
\KwOut{$w$, the attribute importance vector}
 $w = [0, 0, \ldots, 0] \in \mathbf{R}^d$\;
 \For{$t_i \in IF$}{
    node = root($t_i$)\;
    \While{not node.isLeaf()}{
    f = node$.feature$\;
    value = node$.value$\;
    
    \eIf{$x[f] < value$}{
        node = node$.leftChild$\;
    }{
        node = node$.rightChild$\;
    }
    $score = \log_2 \left( \frac{|node.parent|}{|node|} \right) - 1$\;
    $w[f] = w[f] + score$\;
    }
 }
 \Return{w}
 \caption{Computing the attribute importances}
 \label{alg:feature_importance}
\end{algorithm}

\subsection{Assigning Weights to Node Splits}
\label{subsec:weights}
% Weighting scheme captures how much we can shorten the path versus always splitting in half.
% We take into account which node actually contributed to the length of the path (<--> DIFFI method, looks at the full path)
A key contribution of our paper is the definition of an appropriate weighting scheme to evaluate the value of each node split in an isolation tree to the anomaly score of an example traversing the tree. It is based on the intuitions outlined in Section~\ref{subsec:intuitions}. Concretely, the weighting scheme needs to assign a higher value to the split attribute in a node when:
\begin{enumerate}
    \item The split attribute shortens the path length of an example traversing the isolation tree through an imbalanced split.
    \item The split attribute assigns the example to the smaller child node when there is an imbalanced split.
    \item The split attribute occurs in a node with a larger node size. These are the nodes closer to the root and they are more likely to isolate anomalous examples.
\end{enumerate}

% \vincent{blue}{What is the cornerstone? We should be explicit: \emph{Our main insight is that not every split is equally informative in deciding an example's path.}}
% and requires an appropriate weighting scheme to reward splits (and consequently, the chosen attributes for those splits) that help in isolation of the anomalous points of interest. 
% \begin{enumerate}
%     \item Higher weights to imbalanced splits (i.e., splits with $\frac{|node|}{|node.parent|} << \frac{1}{2}$). Higher the imbalance, higher the assigned weight.
%     \item Higher weights to splits involving higher node sizes $|node.parent|$.
%     \item Higher weights to splits that allocate anomalous instance $x$ to the smaller child node of the two. Smaller the allocated child node is to the other, higher the assigned weight.
% \end{enumerate}

\textbf{Balanced Splits}
Before deriving the weight score, let us briefly consider how a \emph{binary search tree} (BST) would split a dataset~\cite{cormen2009introduction}.
The authors of the original \textsc{iForest} paper also identified the structural equivalence between an isolation tree and a binary search tree~\cite{Liu2008-bj}. 
%Indeed, an unsuccessful search in a BST corresponds to the termination of a node in an isolation tree.

Consider a BST where all splits are balanced, with a dataset size of 256 examples. Each node split results in two child nodes of equal sizes, until each example gets isolated to its own leaf node of size 1 at a path length of $log_2(256)=8$.
%For example, let us consider an isolation tree where all splits are balanced (no anomalous point), with a sample of 256 points. Each node split results in children nodes of equal sizes, until each instance gets isolated to leaf nodes of size 1 at a path length of $log_2(256)=8$. 
After the first split, it requires an additional $log_2(128)=7$ splits - or equivalently, $(path length - 1)$ to isolate the example from the second level. In other words, every recursive balanced split takes us \emph{1 unit step} closer to the isolation of the example.

\medskip
\textbf{Imbalanced Splits}
In an isolation tree, the random selection of both split attribute and split value entail that anomalous examples get isolated at shorter path lengths. Hence, the usefulness of selecting a specific attribute to split on depends on how much it contributes to isolating the example.
Hence, our weighting scheme should quantify and capture precisely how much a split reduces the path length of an example compared to its path length in a BST.
%Every recursive balanced split takes us \textbf{1 unit step} closer to the isolation of the instance, given by $|node.parent| - |node|$. Anomalous points get isolated at shorter path lengths on average if anomalous features are chosen for splitting. \textbf{Higher the abnormality, shorter the path length it takes for the point to get isolated}. The usefulness of a specific node split corresponds to be how quickly an anomalous point can get isolated. 

An imbalanced split where the anomalous point gets allocated to the child node with lower size, should assist by \emph{more than 1 unit step} to this example's isolation. For a given example $x$, we compute the weight as:
\begin{align}
\label{eq:score}
    score &= \log_2 \left( \frac{|node(x).parent|}{|node(x)|} \right) - 1
\end{align}
where $node(x)$ is the node example $x$ ends up in after the split.
We subtract $1$ because this is simply the value obtained when the split is perfectly balanced: \begin{align*}
\log_2 \left( \frac{|node(x).parent|}{\frac{|node(x).parent|}{2}} \right) = \log_2(2) = 1
\end{align*}

% \textbf{Larger the assist by a node split, higher the assigned weight}. Using these intuitions, we define the following weighting scheme, which assigns a real-valued number to a split defined by a parent node $\nu$ and a child node $\nu_c$ that contains the anomalous point:

% \begin{equation} \label{eqn:weightingscheme}
% \begin{split}
% & W(node.parent \Rightarrow node) = \frac{Assist_{actual}}{Assist_{prototype}} \\
%  &= \frac{log_2(|node.parent|) - log_2(|node|)}{log_2(|node.parent|) - log_2(\frac{|node.parent|}{2})} \\
%  &= log_2(\frac{|node.parent|}{|node|})
% \end{split}
% \end{equation}

% With this equation, we can plot the output of the weighting scheme (normalized to lie in $[0, 1]$ for different split balances in Figure \ref{fig:bstweightingscheme}.

% \begin{figure}[!h]
%     \centering
%     \includegraphics[width=0.4\textwidth]{Images/bstweighting.png}
%     \caption{Relationship between W and $|node|$ with $|node.parent|=256$}
%     \label{fig:bstweightingscheme}
% \end{figure}
% \vincent{blue}{Probably drop the figure or compress it.}

This weighting scheme rewards imbalanced splits that allocate the anomalous point in question to the smaller child node. A perfectly balanced split (for instance $\{|node.parent|=256 \Rightarrow |node|=128\}$) gets a null reward: $\log_2\frac{256}{128} - 1 = 0$.
The worst possible split (for instance $\{|node.parent|=256 \Rightarrow |node|=255\}$) gets the lowest weight value of $log_2\frac{256}{255}-1=-0.99$.
The best split (for instance, $\{|node.parent|=256 \Rightarrow |node|=1\}$) gets the highest weight value $log_2\frac{256}{1}-1=7$.
It is clear that the score of Equation~\ref{eq:score} varies between $-1$ and $7$.

\subsection{Time Complexity}
\label{subsec:timecomplex}
Our method iterates over every tree in an \textsc{iForest}. For each tree, it follows the path leading to the example $x$, and computes the ratio between each node size and its child size. As described in the original \textsc{iForest} paper, the expected length of that path is $\log_2(|S|)$, with $|S|$ the sample size that is used to fit every tree. Hence, the number of operations is in $O(\log_2(|S|)*T)$. Our method is linear in the number of trees, and logarithmic with respect to the sample size.

%\textcolor{red}{[FILL IN]}

\section{Related Work}
\label{subsec:relatedwork}

Model explainability can be \emph{global} or \emph{local}. Global explainability methods seek to demystify the entire machine learning model, while local methods only aim to explain a model's individual predictions. We are interested in local explainability as we seek to generate an explanation for the anomaly score of each example separately.
Local explainability methods can either be model-agnostic or designed on a model-by-model basis.

\subsection{Model-Agnostic Methods}
Two popular model-agnostic methods are \textsc{Lime}~\cite{Ribeiro2016-mt} and \textsc{Shap} (Shapley values)~\cite{Lundberg2017-ae}.
\textsc{Lime} uses locally interpretable white-box surrogate models to approximate the decision boundaries around an example of interest. It can explain the output of different classifiers, including - theoretically at least - anomaly detectors. However, \textsc{Lime} often produces unstable explanations that are sensitive to the chosen hyperparameters~\cite{alvarez2018robustness}.

% \textbf{Locally Interpretable Model-Agnostic Explanations} (LIME) is a SOTA model-agnostic interpretability method, that uses locally interpretable white-box surrogate models to approximate the decision boundaries around the point of interest \cite{Ribeiro2016-mt}. The local explanation is then provided by analyzing the surrogate white-box model. LIME has successfully been used to explain the output of different classifiers, on different data types (tabular, text and images)~\cite{Ribeiro2016-mt}. However, it has been shown that LIME produces unstable explanations \cite{alvarez2018robustness} and that its hyper-parameters can have a large influence on the produced explanation.

\textsc{Shap} provides both local and global explanations. It assigns a score to each attribute of an example by looking at how the prediction for this example changes when considering only a subset of the attributes. As such, \textsc{Shap} is able to compute the impact of each feature on a prediction. Although it is often consistent with human explanations~\cite{Lundberg2017-ae}, it suffers from instability~\cite{alvarez2018robustness}.
In contrast to both \textsc{Shap} and \textsc{Lime}, our method uses the trees learned by the \textsc{iForest} to provide the explanations. Our method does not probe the \textsc{iForest} by either sampling similar data points or masking features, but uses the model that is actually learned to provide explanations.

% \textbf{SHAP} is another model-agnostic interpretability method that can be used for both local explanations and global model interpretability \cite{Lundberg2017-ae}. The method is based on Shapley values that originated in the field of coalitional game theory.
% %in the late 50s \cite{Shapley1953-dd} and described the optimal way to allocate the gain in payout to the game players.
% %The same framework is employed in Machine Learning wherein
% SHAP assigns a score to each feature by looking at how a prediction changes when considering only a subset of the original features. By looking at all feature combinations, SHAP is able to compute the impact of each feature on a given prediction. SHAP has been shown to be more consistent with human explanations~\cite{Lundberg2017-ae}. It however suffers from unstability \cite{alvarez2018robustness}.
%The prediction task is treated as the game being played, feature/attribute values of the point of interest as the different players in the game, the average model output as average game payout and finally, the residual between the model output for the point of interest \& average model output as the \textit{gain in payout} that needs to be allocated to each of the feature values (players). In fact, SHAP is the only local explanation method with a solid theoretical background and a strong game-theoretic foundation \cite{Strumbelj2014-fa}.
% \vincent{blue}{This paragraph is too long + does not clearly explain how SHAP values are connected to interpretability.}

\subsection{Methods Specific to Anomaly Detection}
\label{subsec:diffi_rel_work}

Few explainability methods for anomaly detection algorithms exist.
Most related to our work is the post-hoc \textsc{Diffi} method~\cite{Carletti2020-hp}. It computes the importance of each attribute in the \textsc{iForest} algorithm based on the tree depth.
In contrast to \textsc{Diffi}, our method also takes into account the imbalance of a node split in each tree of the \textsc{iForest} ensemble when deriving an explanation.

In local-\textsc{Diffi}, the attribute importances are also computed in an additive manner by assigning a weight to each split node. However, the node weight is computed as $\frac{1}{h_{t_i}(x)}-\frac{1}{h_{max}}$ where $h_{t_i}(x)$ represents the height of the leaf node of $x$, and $h_{max}$ is the expected leaf depth of balanced splits: $\log_2(|root(t_i)|)$. This weighting scheme therefore assigns a constant weight to all nodes on a given path, based on the depth of that path. This contrasts with our method which assigns a weight to each node based on the imbalance of this specific node. In other words, the local-DIFFI weighting scheme does not capture the intuition that split nodes that shorten the path length of an example should be rewarded. Our method does take into account the contribution of each node to the path length of an example. This is important because an attribute that is instrumental to isolating the example probably also provides an explanation why the example is anomalous.

\begin{figure*}[t]
    \centering
    \begin{subfigure}{0.25\textwidth}
        \includegraphics[width=\textwidth]{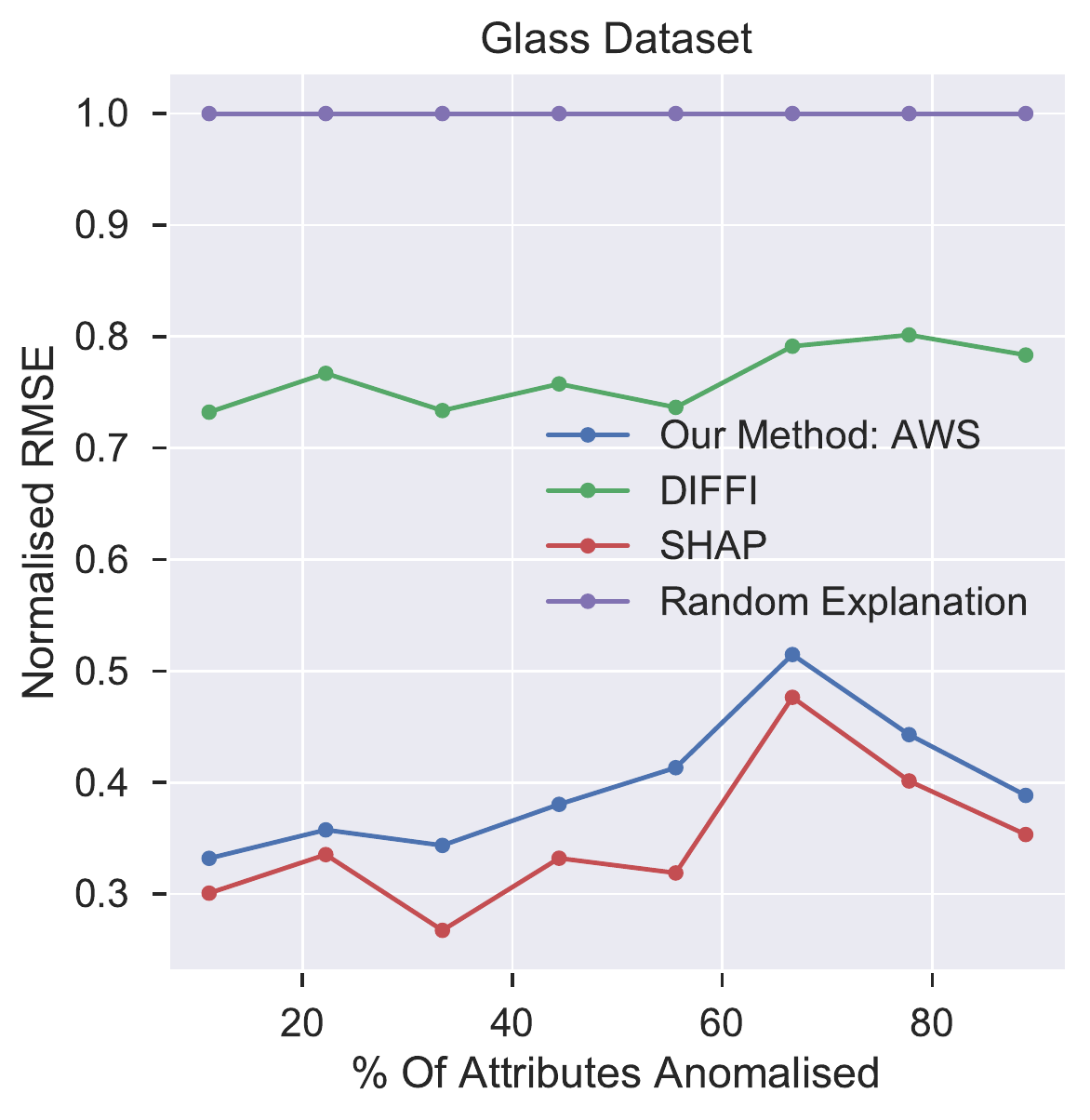}
        \label{fig:a}
    \end{subfigure} \hspace{8mm}%
    \begin{subfigure}{0.25\textwidth}   
        \includegraphics[width=\textwidth]{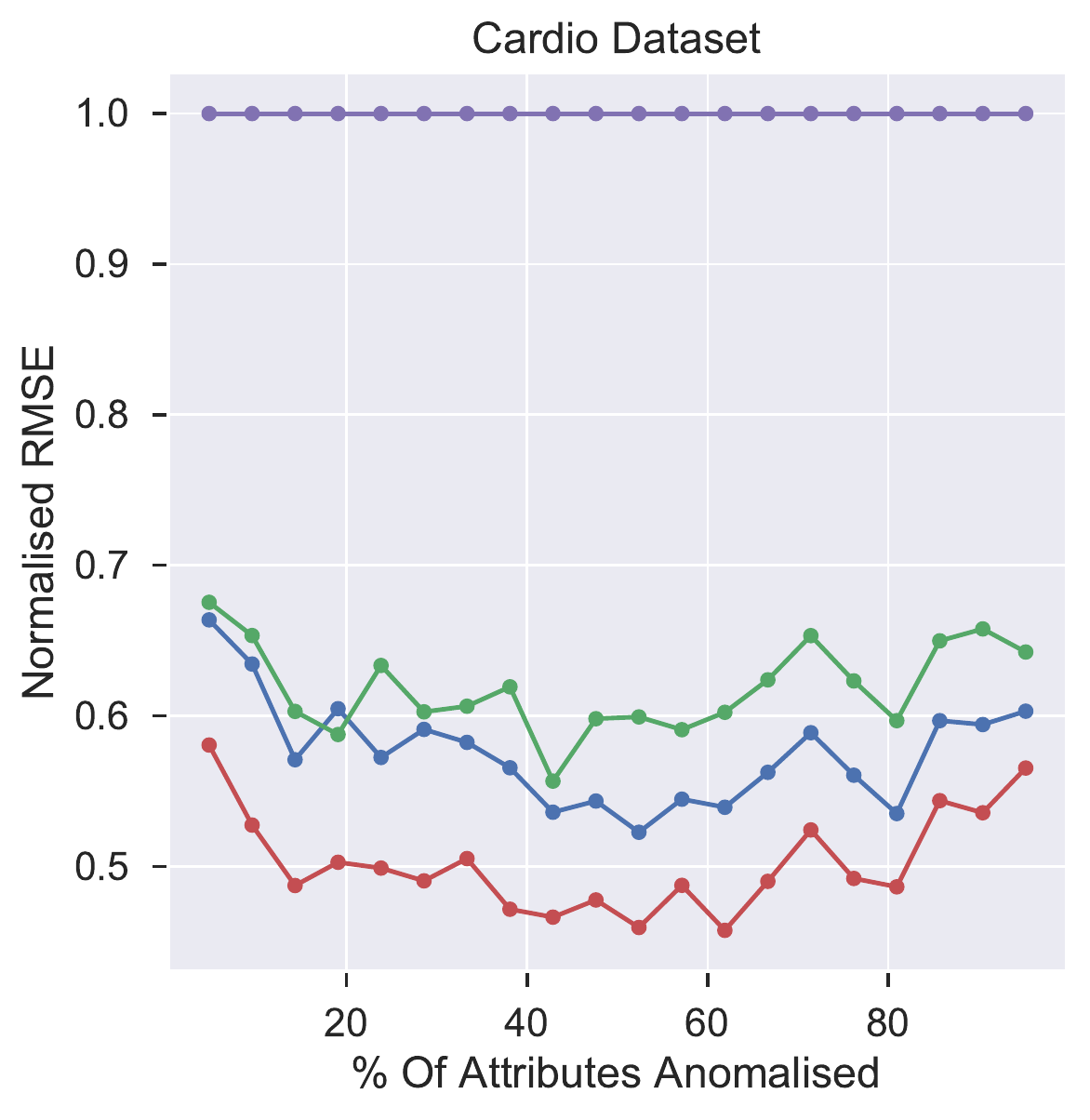}
        \label{fig:b}    
    \end{subfigure} \hspace{8mm}%
    \begin{subfigure}{0.25\textwidth}    
        \includegraphics[width=\textwidth]{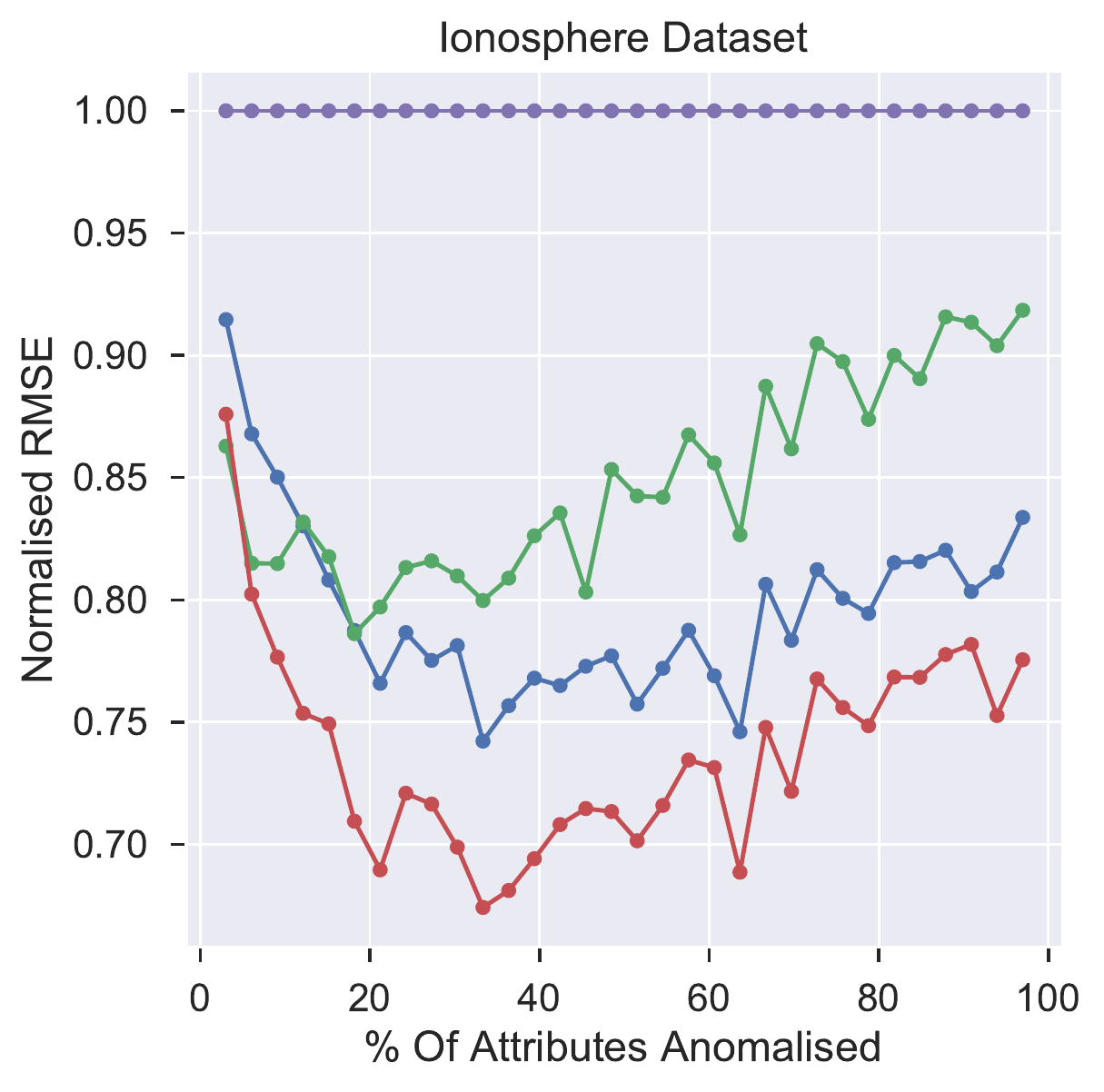}
        \label{fig:b}    
    \end{subfigure} 
    
    \begin{subfigure}{0.25\textwidth}
        \includegraphics[width=\textwidth]{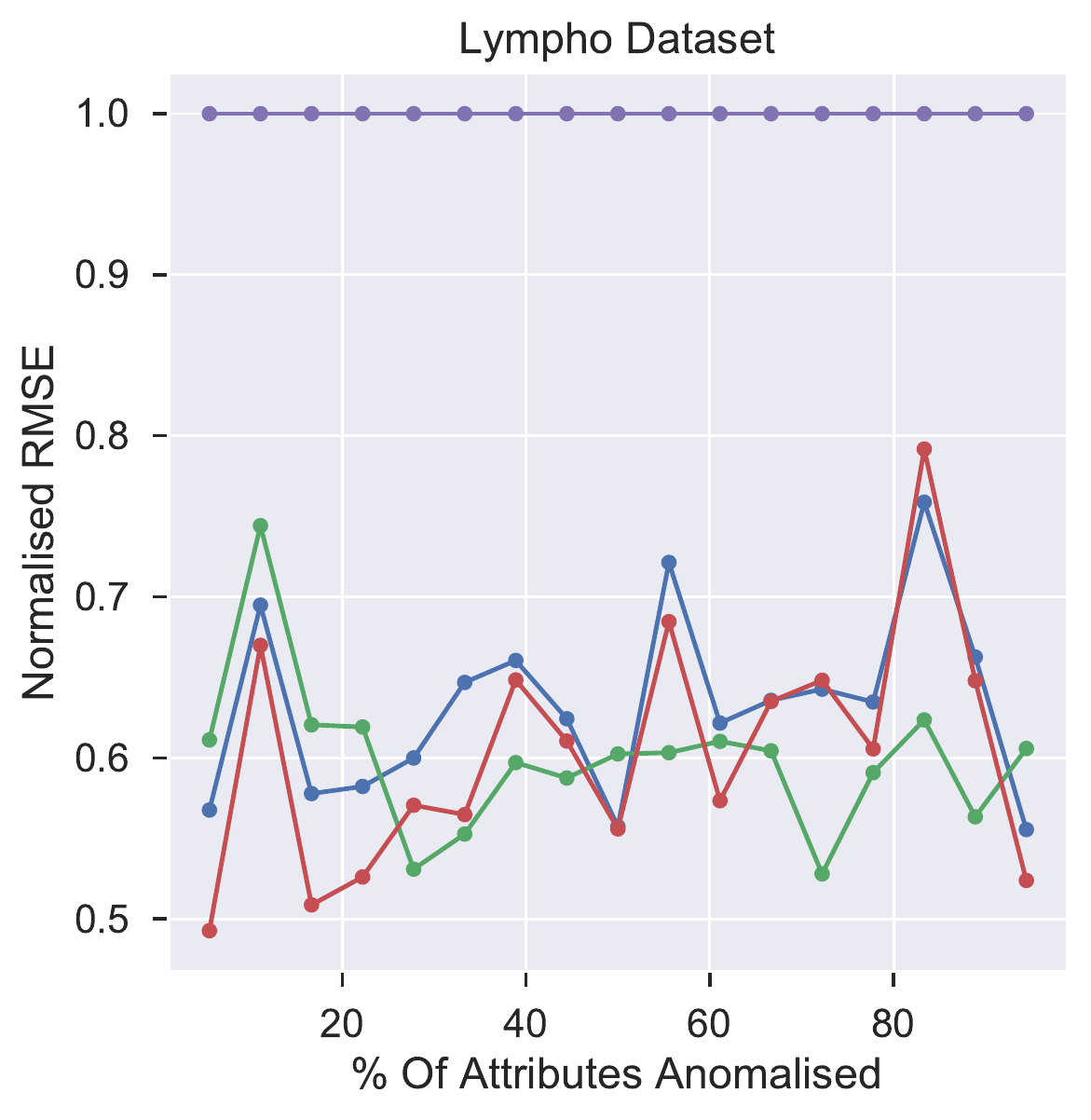}
        \label{fig:a}
    \end{subfigure} \hspace{8mm}%
    \begin{subfigure}{0.25\textwidth}    
        \includegraphics[width=\textwidth]{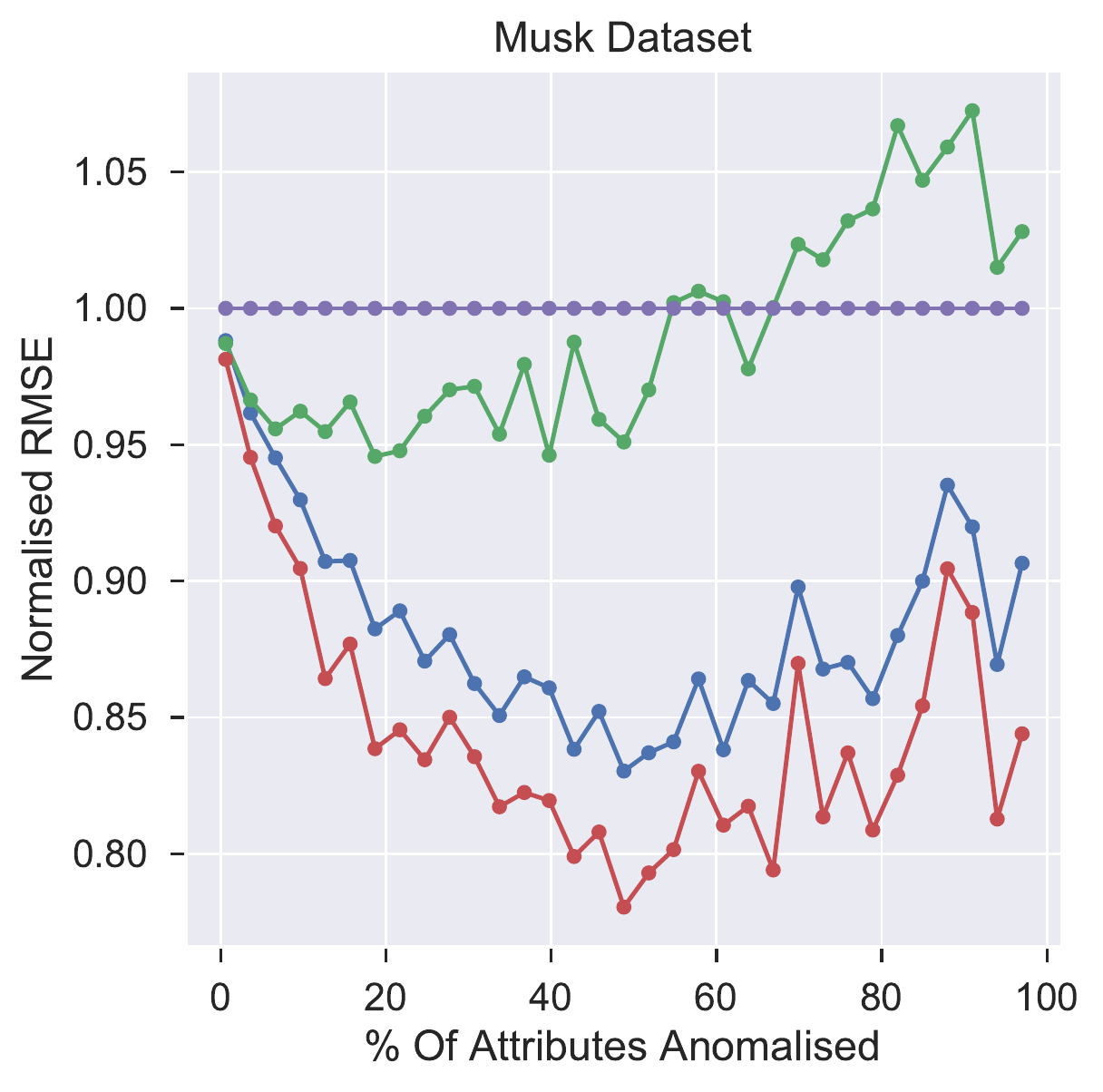}
        \label{fig:b}    
    \end{subfigure} \hspace{8mm}%
    \begin{subfigure}{0.25\textwidth}    
        \includegraphics[width=\textwidth]{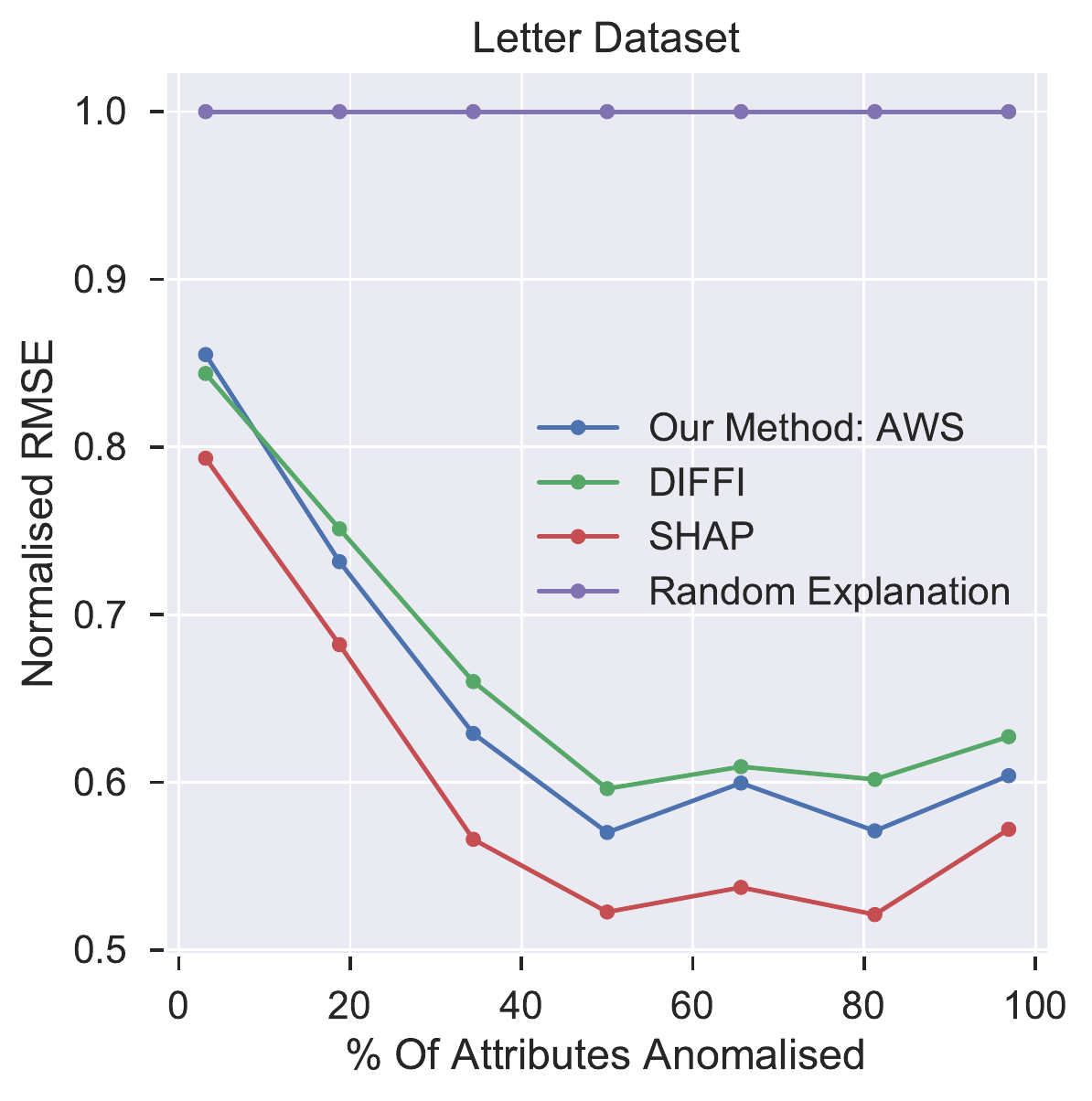}
        \label{fig:b}    
    \end{subfigure} 
    \caption{The plots show the evolution of the RMSE against the percentage ($m$) of \emph{anomalized} attributes for each of the six real-world datasets. RMSE are normalized using the random baseline. The plots clearly show that our method performs better than or on par with \textsc{Diffi} for each dataset. It also matches the performance of \textsc{Shap} (or is very close to it) in each dataset.}
    \label{fig:accuracy}
\end{figure*}

\begin{figure*}[t]
    \centering
    \begin{subfigure}{0.25\textwidth}
        \includegraphics[width=\textwidth]{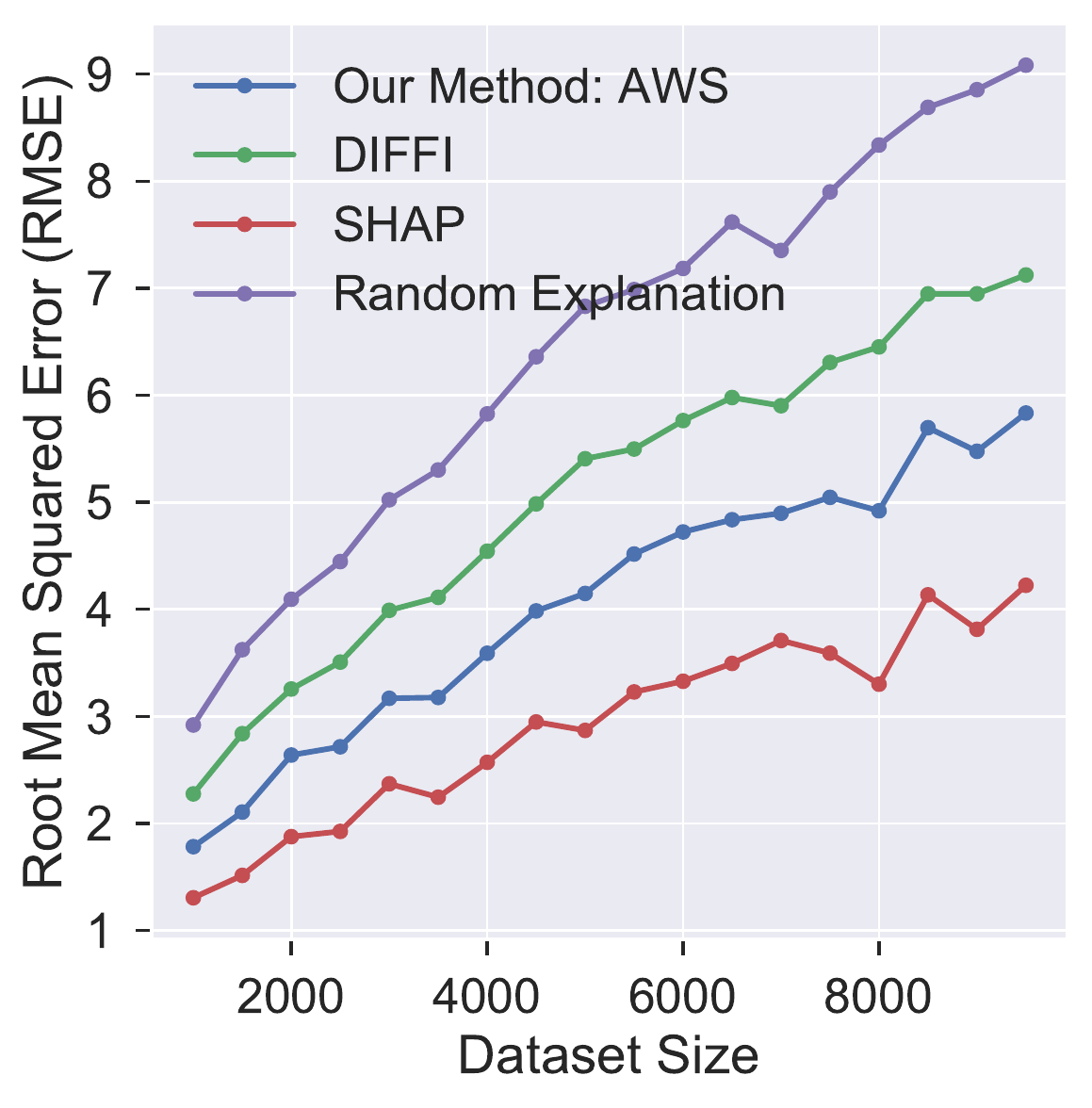}
        \subcaption{RMSE vs dataset size.}
    \end{subfigure} \hspace{8mm}%
    \begin{subfigure}{0.25\textwidth}    
        \includegraphics[width=\textwidth]{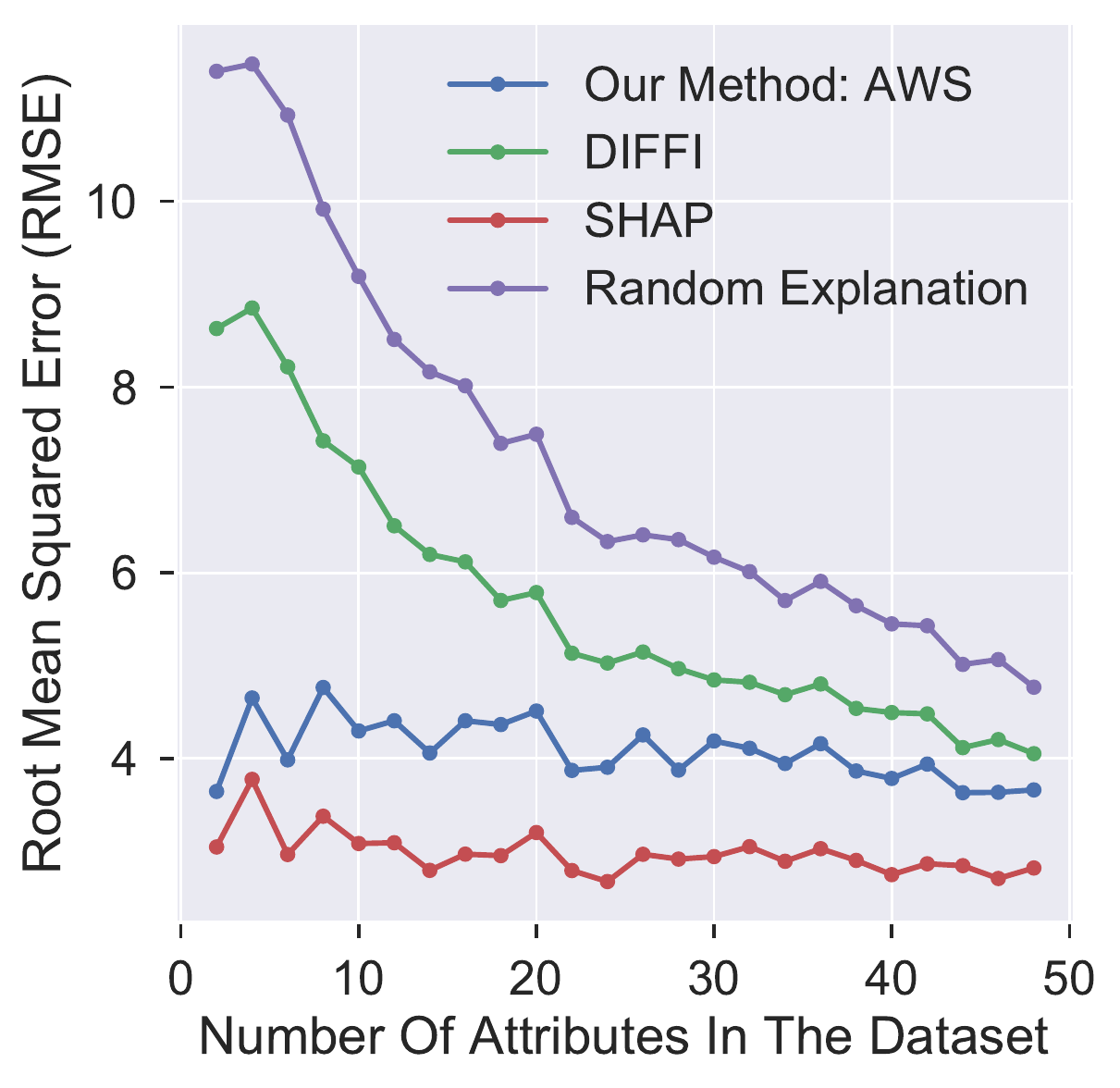}
        \subcaption{RMSE vs dimensionality.}
    \end{subfigure} \hspace{8mm}%
    \begin{subfigure}{0.25\textwidth}    
        \includegraphics[width=\textwidth]{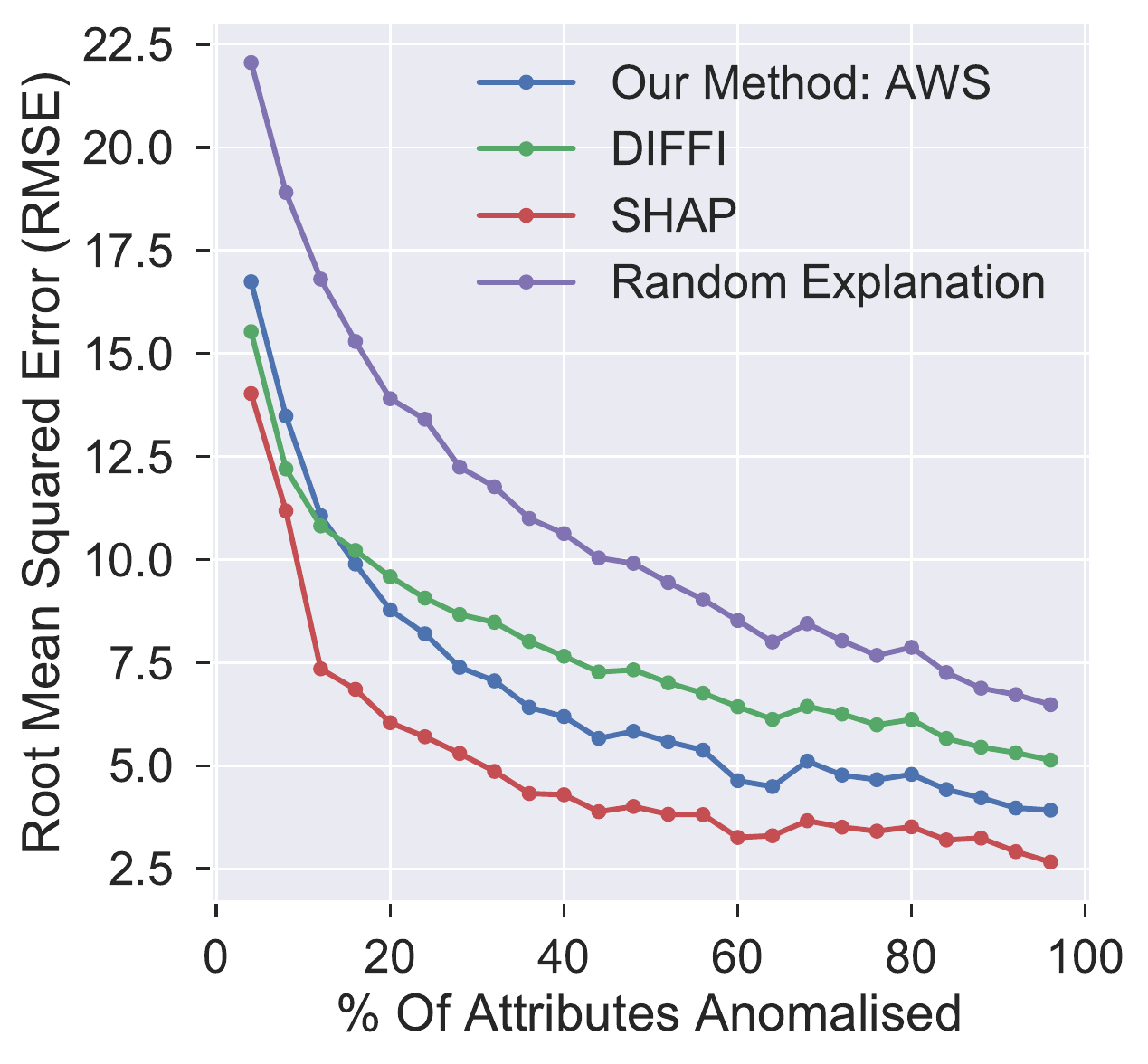}
        \subcaption{RMSE vs \% anomalized attr.}
    \end{subfigure} 
    \caption{The plots show the results of the experiments on the synthetic data.}
    \label{fig:synthethic}
\end{figure*}

\section{Experiments}

In this section, we try to answer the following questions:
\begin{description}
\itemsep0em 
    \item[\textbf{Q1:}] Does our method provide accurate local explanations for examples flagged as anomalies?
    \item[\textbf{Q2:}] Is our method more efficient than comparable state-of-the-art methods?
    \item[\textbf{Q3:}] What are the key differences between local-\textsc{Diffi} and our method?
\end{description}

\medskip

\textbf{Methods}
We compare the following methods: the model-agnostic local \textsc{Shap} values~\cite{Lundberg2017-ae}, the local-\textsc{Diffi} method designed for the \textsc{iForest} algorithm~\cite{Carletti2020-hp}, and our method as described in Section~\ref{sec:method}.
The methods do not have specific hyperparameters to set. Each method explains the output of the same constructed \textsc{iForest} with $T = 100$ and $\psi = 256$.

% For each method, we set its hyperparameters to the recommended values of the original papers. \textcolor{red}{[FILL IN]} \kartha{green}{No hyperparameters for all 3 methods. They require just the trained model.}
\medskip

\textbf{Data} We run our experiments on both synthetic and real-world data.\footnote{A full implementation of the experiments and methods is provided online: \url{https://github.com/karthajee/AWS-IF-interpretability}}
The synthetic data consist of multiple settings of normally distributed point clusters wherein we systematically vary dataset size (from 1,000 to 10,000), dimensionality (from 2 to 50), and the proportion of attributes randomly chosen examples are made anomalous across (till 100\%).
The six real-world data are listed in Table~\ref{table:datasets}. For the sake of consistency, these are the same datasets as used in the \textsc{Diffi} paper~\cite{Carletti2020-hp}.
%\kartha{green}{Could not complete satellite dataset. System crashed and we do not have time. Replace "satellite" with "glass" - results have already been uploaded and DIFFI paper also used them.}

\begin{table}[t]
\centering
\resizebox{0.35\textwidth}{!}{%
\begin{tabular}{@{}lll@{}}
\toprule
\textbf{Dataset ID} & \textbf{Nr. of examples} & \textbf{Nr. of features} \\ \midrule
\texttt{glass} & 214 & 10 \\
\texttt{cardio} & 1831 & 21 \\
\texttt{ionosphere} & 351 & 33 \\
\texttt{lympho} & 148 & 18 \\
\texttt{musk} & 3062 & 166 \\
\texttt{letter} & 1600 & 32 \\ \bottomrule
\end{tabular}%
}
\caption{Six real-world datasets from the UCI repository~\cite{Dua:2019}.}
\label{table:datasets}
\end{table}

\medskip
\textbf{Experimental Setup}
The fundamental design issue with explainability experiments is the \emph{absence of a ground truth} to evaluate the predictions of the methods.
Hence, we design the following procedure that simulates a ground-truth.
Given a datasets consisting only of normal examples and an \textsc{iForest} trained on these data, we randomly pick $n$ examples with replacement and for each example:
(1) obtain an \emph{initial} explanation vector $w_0$,
(2) randomly select a coalition of $m\%$ of the example's attributes,
(3) for each of those, change its value to the $3*$max value of that particular attribute's full range over all the data,
(4) obtain a \emph{new} explanation vector $w_{new}$ for the modified example,
(5) compute the change between the initial and new explanations as $w = w_0 - w_{new}$ and normalize $w$. 
We now expect the change in explanation to occur only for the coalition of attributes that we changed, captured by the expected explanation vector $w_e$.
This allows us to compute the \emph{root mean squared error} (RMSE) between vectors $w$ and $w_e$. For instance, if we change the value of only one attribute in steps (2) and (3), we would expect $100\%$ of the change in the explanation to occur for this particular attribute, reflected in vector $w$.
Finally, we average the errors over all $n$ randomly picked examples.
%We repeat this experiment times and report the averaged results for each method. 
We also compute the RMSE for a random explanation vector.

\begin{table*}[]
    \centering
    \resizebox{\textwidth}{!}{%
    \begin{tabularx}{1.2\textwidth}{l *{12}{Y}}
    \toprule
     & \multicolumn{6}{c}{Execution time per example (sec) with $m = 10\%$}  
     & \multicolumn{6}{c}{Execution time per example (sec) with $m = 100\%$}\\
    \cmidrule(lr){2-7} \cmidrule(l){8-13}
    \textbf{Method}  & \texttt{glass} & \texttt{cardio} & \texttt{iono} & \texttt{lympho} & \texttt{musk} & \texttt{letter} & \texttt{glass} & \texttt{cardio} & \texttt{iono} & \texttt{lympho} & \texttt{musk} & \texttt{letter} \\
    \midrule
    \textbf{Our method} & \textbf{0.027} & 0.040 & 0.041 & \textbf{0.040} & 0.040 & \textbf{0.029} & \textbf{0.026} & 0.039 & 0.040 & \textbf{0.039} & 0.039 & \textbf{0.028} \\
    \textsc{Diffi} & 0.029 & \textbf{0.037} & \textbf{0.038} & 0.045 & \textbf{0.039} & 0.030 & 0.030 & \textbf{0.037} & \textbf{0.038} & 0.040 & \textbf{0.037} & 0.029 \\
    \textsc{Shap} & 0.255 & 0.213 & 0.230 & 0.305 & 0.244 & 0.274 & 0.269 & 0.196 & 0.243 & 0.264 & 0.216 & 0.251 \\
    \bottomrule
    \end{tabularx}%
    }
    \caption{The table shows the execution time per example for different percentages of \emph{anomalized} attributes for each of the six real-world datasets. Our method is an order of magnitude faster than \textsc{Shap}. It performs on par with \textsc{Diffi}. The percentage of anomalized attributes does not influence the execution time (i.e., the inference of an example's explanation vector).}
    \label{tab:execution_time}
\end{table*}
% \textcolor{green}{We measure execution time PER SAMPLE - That needs to be mentioned in the table}

\subsection{Experimental Results}

\textbf{Q1: Accuracy of the Explanation Methods}
In this experiment, we evaluate the accuracy of our explanation method. The latter is defined as the RMSE between the expected explanation $w_e$ and the provided explanation $w$. For instance, if an example is \emph{anomalized} among $k$ attributes, its ideal explanation vector has value $\frac{1}{k}$ for each of these $k$ attributes, and $0$ for the others.
Figure~\ref{fig:accuracy} shows for each of the six real-world datasets how the RMSE values of the compared methods change as a function of $m$. The RMSE values are normalized using the performance of the random baseline.
Our method performs better or equal to local-\textsc{Diffi} on each dataset. Its performance is similar to \textsc{Shap} on each dataset.
% \textcolor{red}{[explain why RMSE decreases?]}

We conduct experiments on synthetic data to measure the impact on accuracy of systematically varying: the dataset size, the dataset dimensionality, and the percentage of anomalized attributes.
Figure~\ref{fig:synthethic}(a) shows the RMSE of the different methods for a varying dataset size.
An increase in dataset size increases the RMSE of each method in a similar fashion.
Figure~\ref{fig:synthethic}(b) shows the RMSE of the different methods for a varying number of attributes. In this experiment, $m = 100\%$. The performance of our method and \textsc{Shap} are not affected by varying dimensionality, in contrast to local-\textsc{Diffi}.
Figure~\ref{fig:synthethic}(c) shows the RMSE of the different methods for a varying number of anomalized attributes. In both Figure~\ref{fig:synthethic}(b) and Figure~\ref{fig:synthethic}(c), the RMSE is decreasing when the number of anomalized attributes is decreasing. The computed RMSE is $||w_e-w||_2 = \sqrt{\sum_{i=1}^{d}(w_e(d) - w(d))^2} = \sqrt{\sum_{i=1}^{d}(w_e(d)^2 + w(d)^2 - 2*w_e(d)*w(d))}$. When the number of anomalized attributes increases, the number of non zero elements in $w_e$ increases, hence it is more likely than $2*w_e(d)*w(d)$ is non zero. Moreover, recall that $w_e$ has value $\frac{1}{k}$ for each anomalized attribute, so $\sum_{i=1}^{d}w_e(d)^2=k*\frac{1}{k^2}=\frac{1}{k}$. Hence, we have RMSE$ = \sqrt{\frac{1}{k} + \sum_{i=1}^{d}(w(d)^2 - 2*w_e(d)*w(d))}$ so the RMSE is expected to resemble the inverse function with respect to the number of anomalized attributes.

% \clement{orange}{For real dataset, doing the explanation shift makes sense. For synthetic dataset, how is it exactly generated? And why can't we look at the explanation vector directly? In a practical setting, users will generate an explanation vector for a point, not look at the shift. hence, we should validate that the vector, not only the shift, makes sense.}

% \textcolor{red}{[FILL IN]}

% \textcolor{green}{This was earlier, when we were running 20 iterations. Now, we systematically vary dimensionality, number of features and \% of anomalised attributes, one at a time. We do not conduct multiple experiments for the same config.}

\medskip
\textbf{Q2: Efficiency of the Explanation Methods}
%We measure the efficiency of the different methods using their execution time during inference (i.e., computing the explanation vector for an example).
Table~\ref{tab:execution_time} shows the execution time during inference (i.e., computing the explanation vector for an example) for each method on the six datasets for different percentages of anomalized attributes.
As both local-\textsc{Diffi} and our method use similar approaches, it is expected that their execution times are comparable. As noted in Section~\ref{subsec:timecomplex}, the complexity of our method depends on the number of trees and the sample size. As both of these variables are kept constant, the execution time of our method also remains constant.
Because it exploits the structure of the \textsc{iForest}, our method is an order of magnitude faster than \textsc{Shap}.

%\clement{orange}{As runtime is constant, should we look at different settings (nb of forest, sample size) and see how the runtime evolves?}

\begin{figure*}[hbt]
    \centering
    \begin{subfigure}{0.23\textwidth}
        \centering
        Our method: IB, A
        \includegraphics[width=\textwidth]{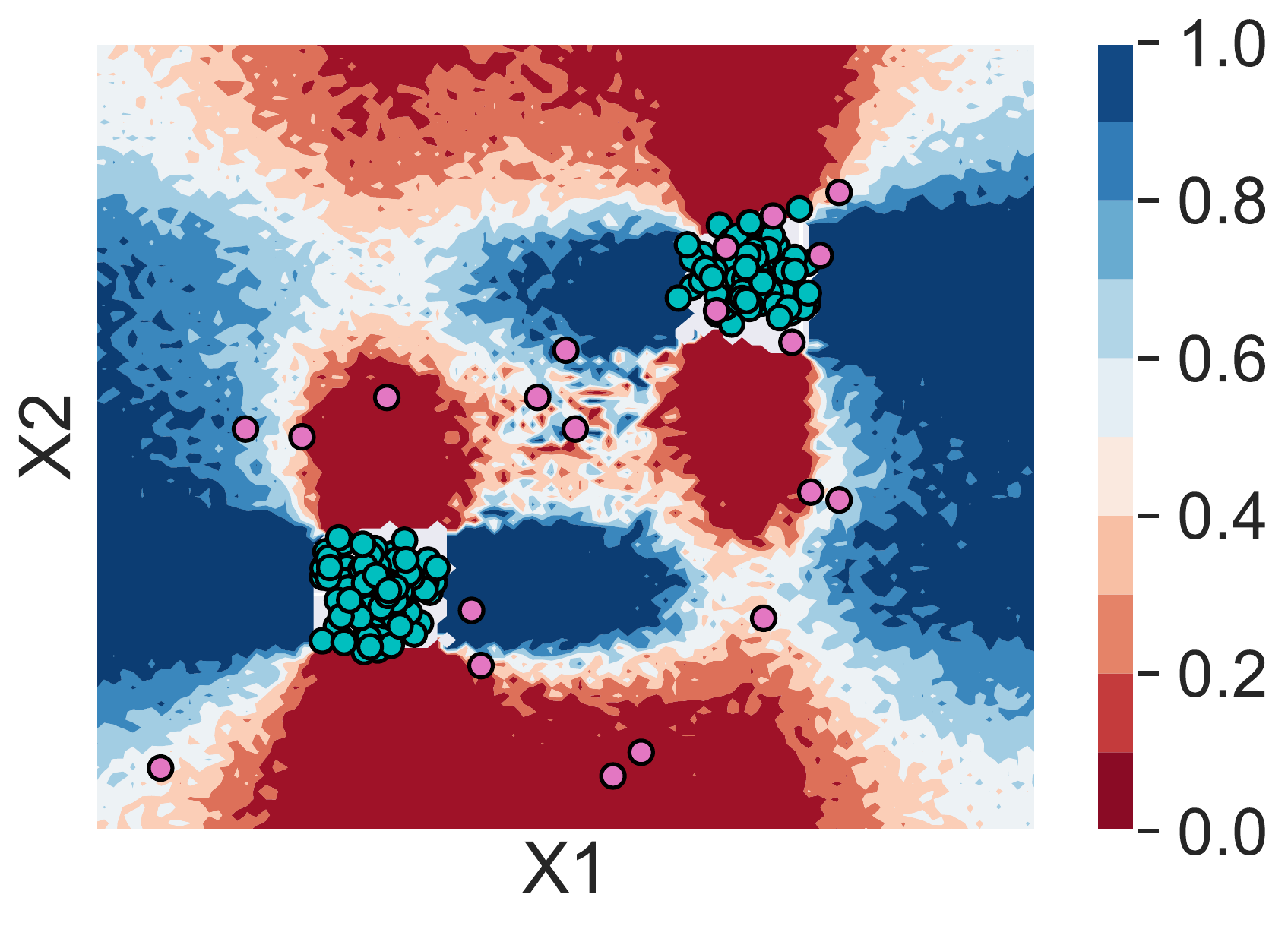}
        \label{fig:a_maps}
    \end{subfigure} %
    \begin{subfigure}{0.23\textwidth}   
        \centering
        Our method: OOB, A
        \includegraphics[width=\textwidth]{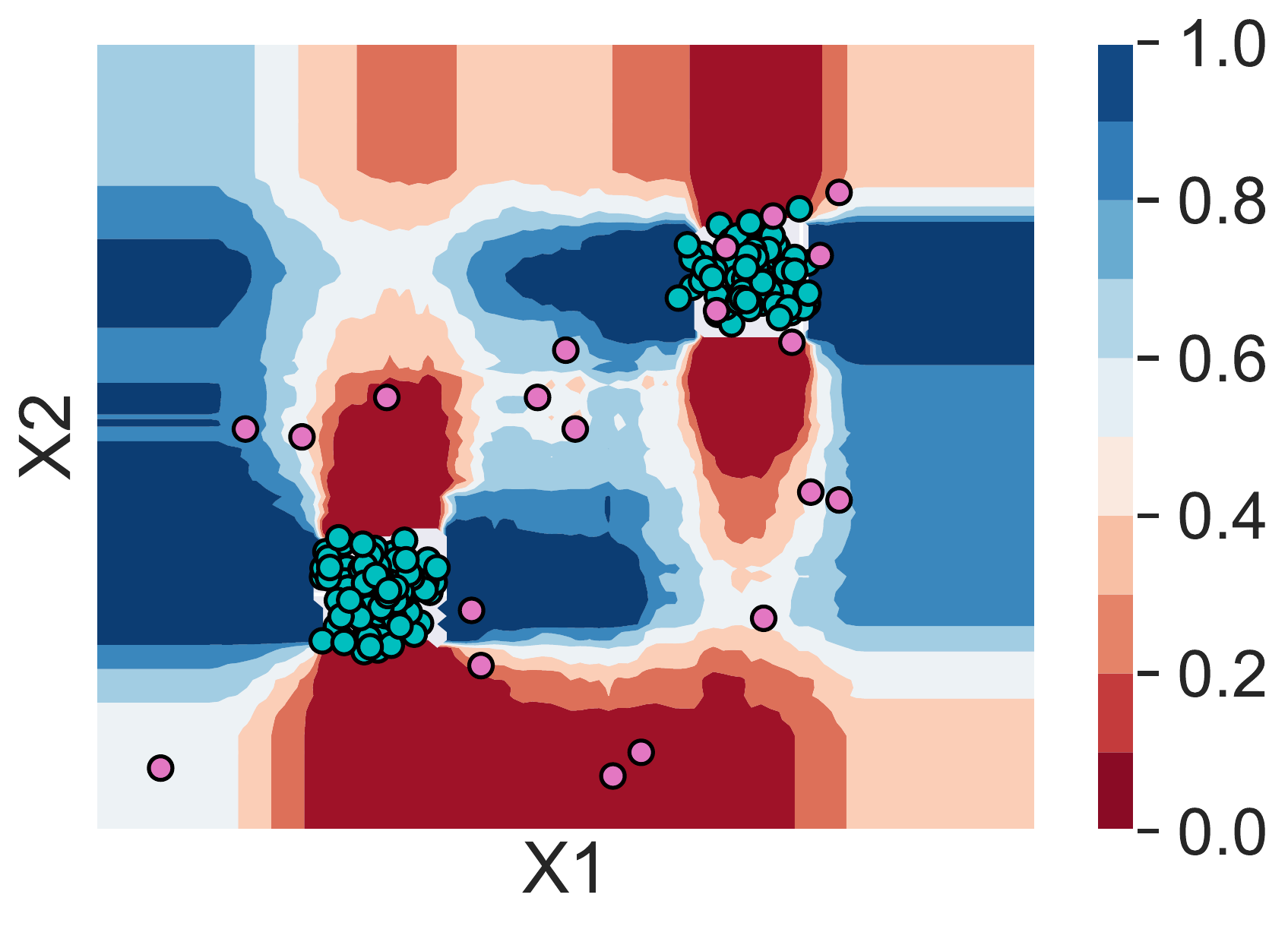}
        \label{fig:b_maps}    
    \end{subfigure} %
    \begin{subfigure}{0.23\textwidth}
        \centering
        Our method: IB, No-A
        \includegraphics[width=\textwidth]{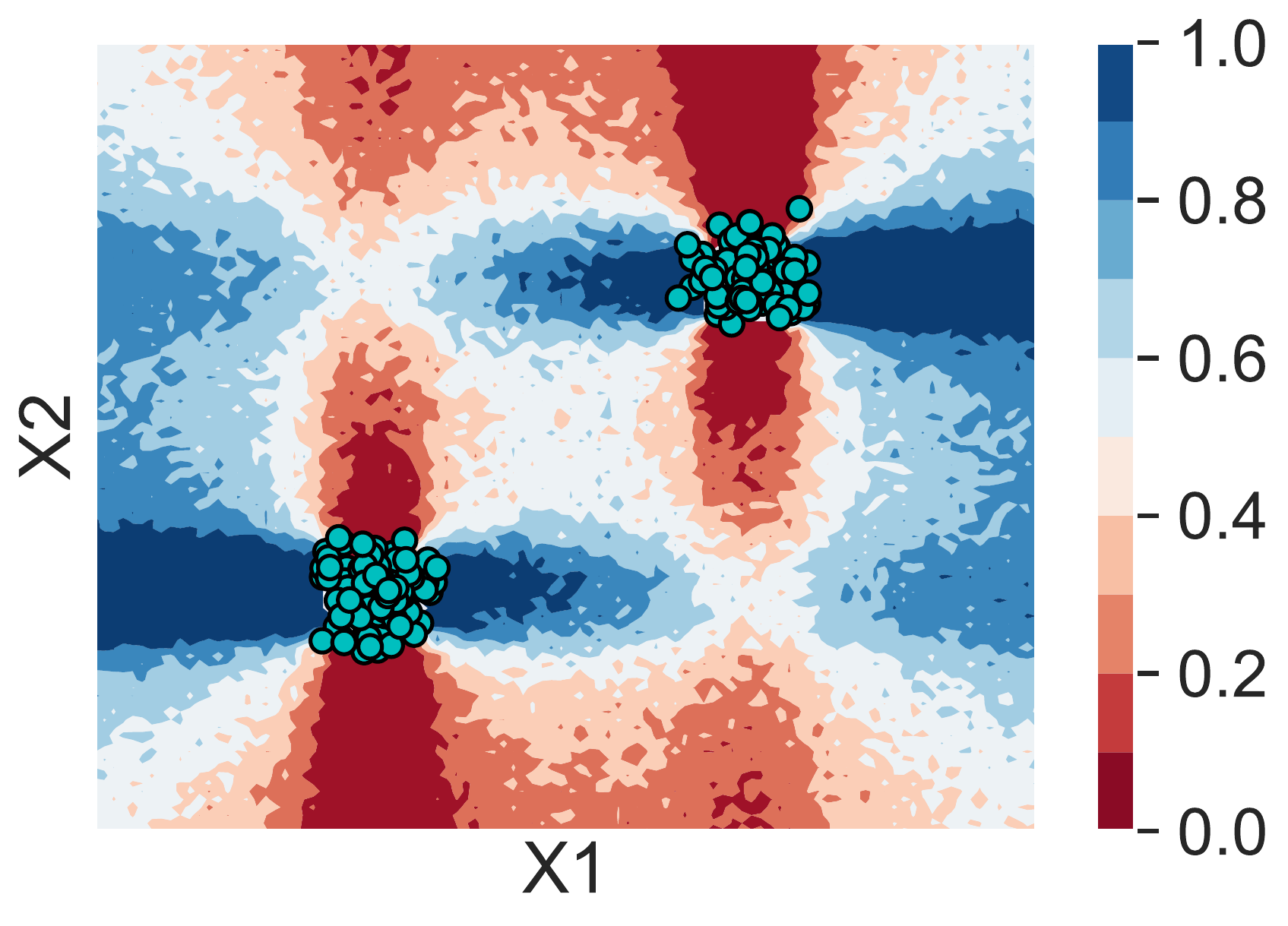}
        \label{fig:c_maps}
    \end{subfigure} %
    \begin{subfigure}{0.23\textwidth}
        \centering
        Our method: OOB, No-A
        \includegraphics[width=\textwidth]{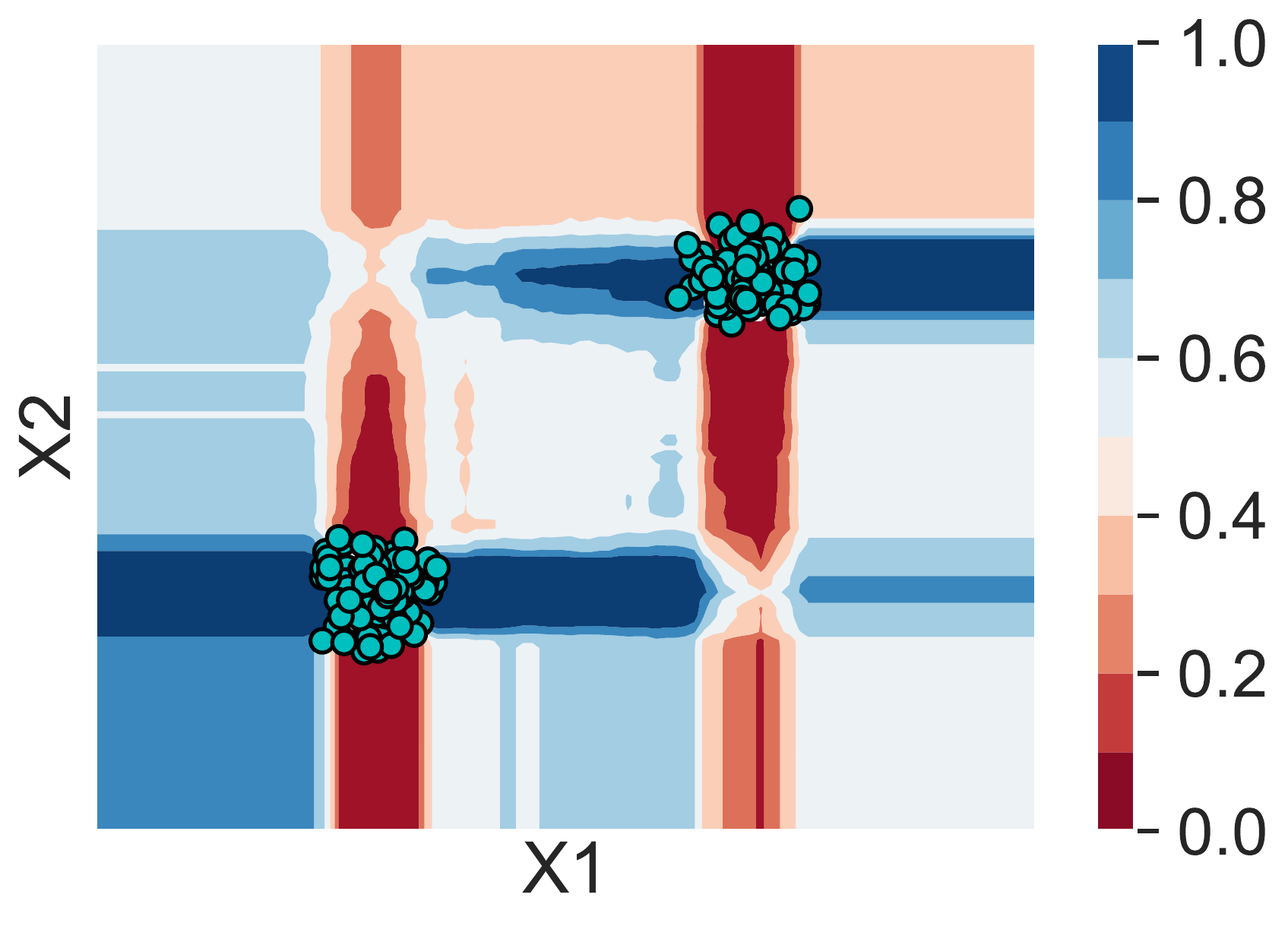}
        \label{fig:d_maps}    
    \end{subfigure} %
    
    \begin{subfigure}{0.23\textwidth}
    \centering
        \textsc{Diffi}: IB, A
        \includegraphics[width=\textwidth]{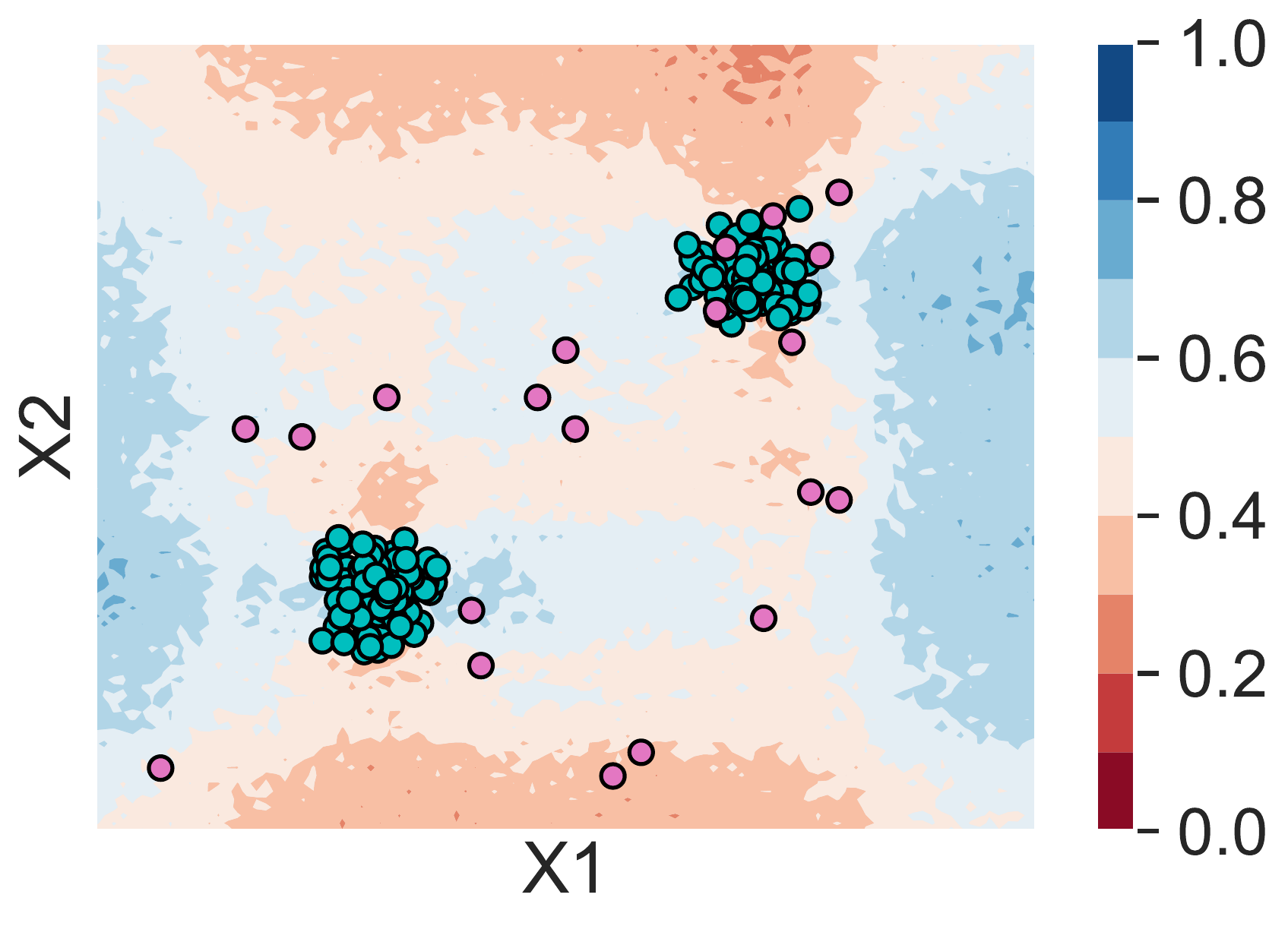}
        \label{fig:e_maps}
    \end{subfigure} %
    \begin{subfigure}{0.23\textwidth}  
    \centering
        \textsc{Diffi}: OOB, A
        \includegraphics[width=\textwidth]{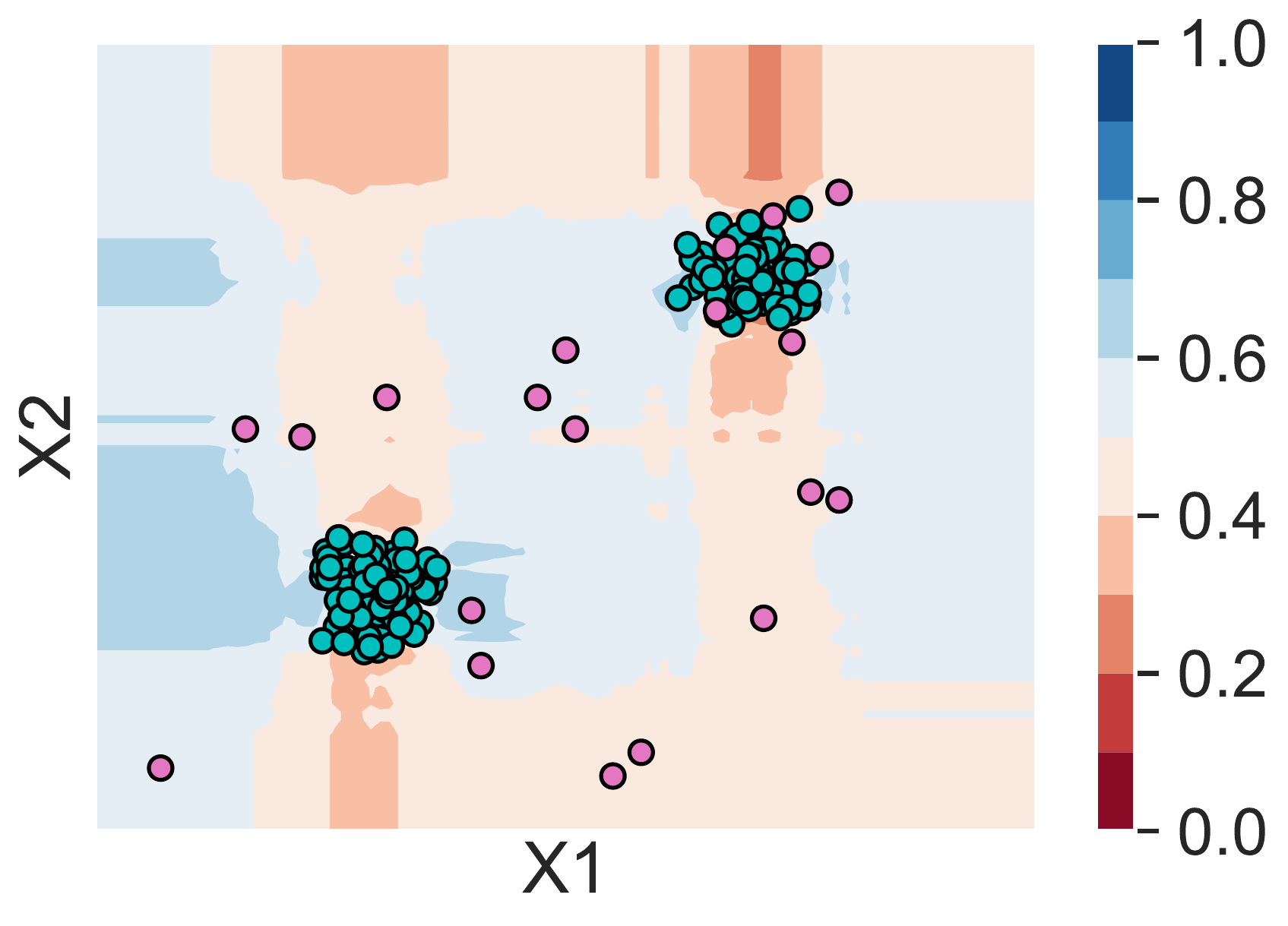}
        \label{fig:f_maps}    
    \end{subfigure} %
    \begin{subfigure}{0.23\textwidth}
     \centering
        \textsc{Diffi}: IB, No-A
        \includegraphics[width=\textwidth]{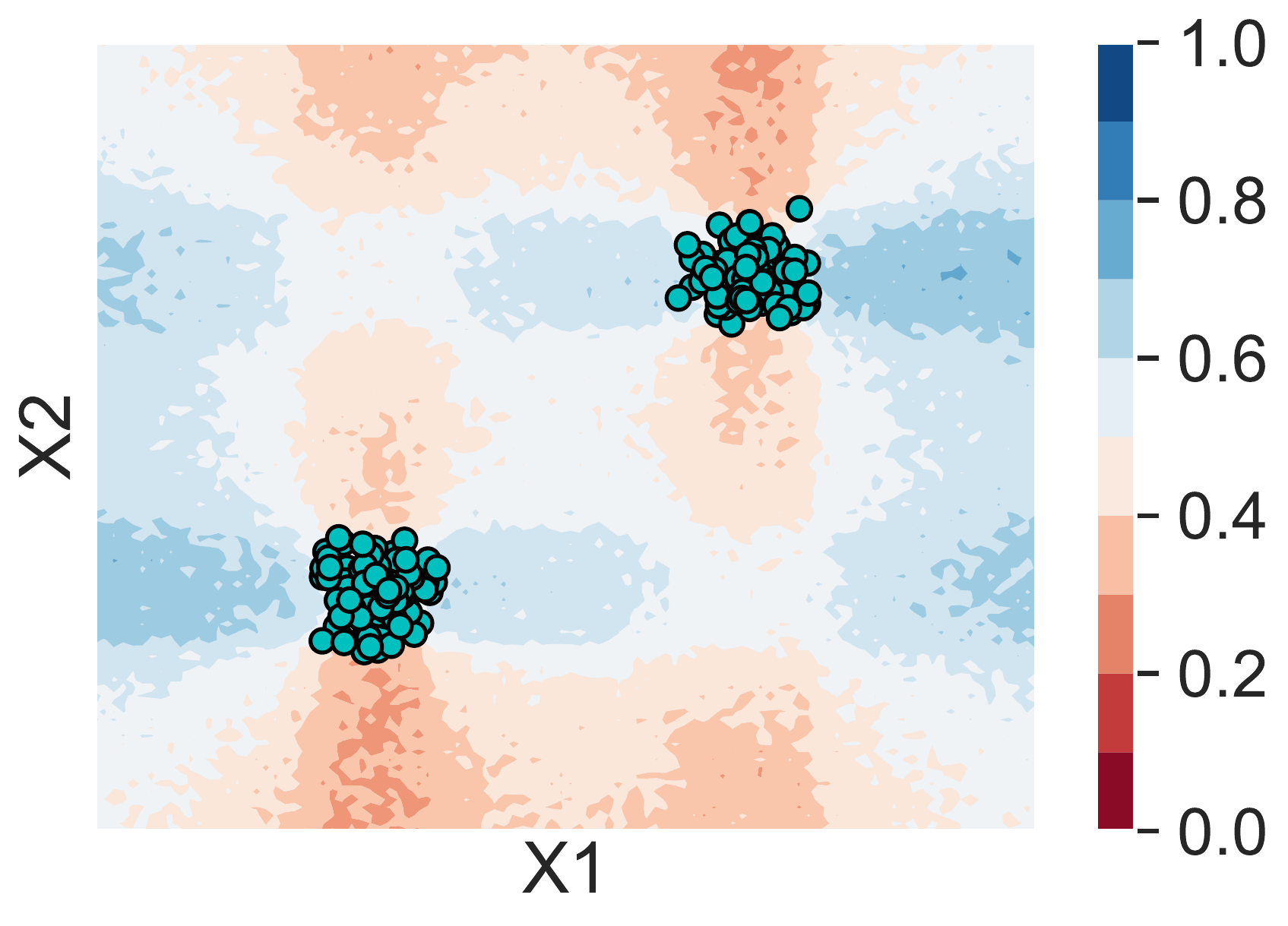}
        \label{fig:g_maps}
    \end{subfigure} %
    \begin{subfigure}{0.23\textwidth} 
    \centering
        \textsc{Diffi}: OOB, No-A
        \includegraphics[width=\textwidth]{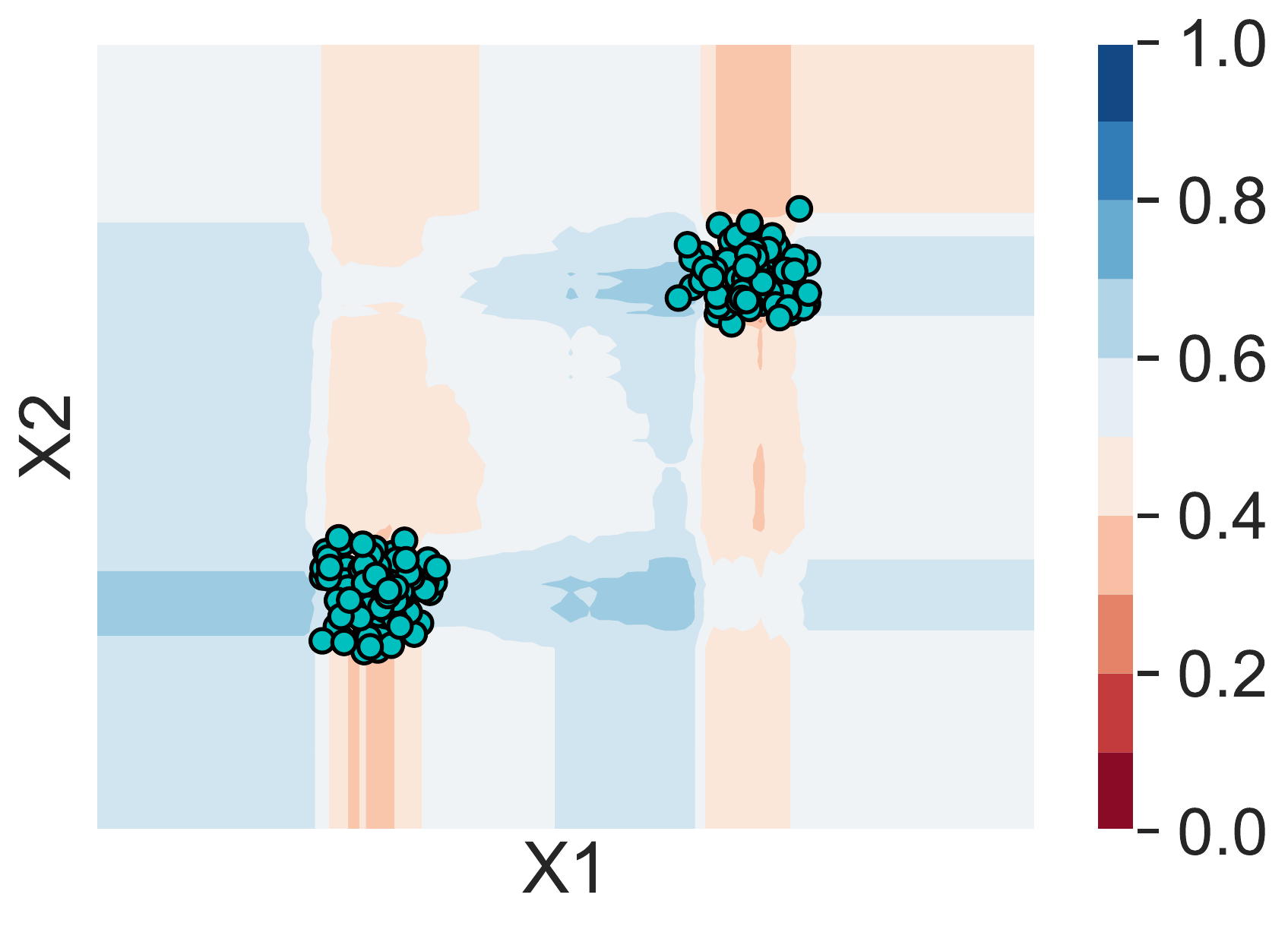}
        \label{fig:h_maps}    
    \end{subfigure} %
    \caption{Heatmaps representing the contribution of the each attribute to the anomaly score (blue = $X1$ and red = $X2$). For each plot, one cluster lies on the bottom left and the other in the top right quadrant of the plot. Explanation vectors are normalised so their values sum to 1. The first row depicts heatmaps of our method, the second row depicts heatmaps of local-\textsc{Diffi}. When the \textsc{iForest} is retrained for each heatmap point, we say that the setting is \textit{in-bag} (IB), as the data point is part of the trained \textsc{iForest}. When the\textsc{iForest} is not retrained, the setting is said to be \textit{out-of-bag} (OOB). When there are anomalies in the training data, the setting is \textit{with anomalies} (A). When there are no anomalies in the training data, the setting is said to be \textit{no anomalies} (No-A). There are therefore a total of 4 settings for each method.}
    \label{fig:heatmaps}
\end{figure*}

\medskip
\textbf{Q3: Key differences with local-\textsc{Diffi}}
As explained in Section \ref{subsec:diffi_rel_work}, our method differs from local-\textsc{Diffi} by defining a new scheme to weight node splits. In local-\textsc{Diffi}, all nodes along a path are allocated a constant weight depending solely on the depth of that path. Our method instead rewards each node based on the imbalance it created towards isolating the considered example. While both methods reward splits that lead to shorter path lengths, our method treats node individually.
We illustrate in a toy example how this key difference impacts the explanation provided by both methods, and argue that our method addresses a shortcoming of local-\textsc{Diffi} weighting scheme.

The toy example is composed of a 2D two-cluster configuration and is depicted in Figure~\ref{fig:heatmaps}. The heatmaps depict the contribution of each attribute ($X1$ and $X2$) to the anomaly score. Blue means $X1$ explains the score, while red means that $X2$ explains the score.
We consider four different settings when computing the heatmaps.
First, the \textsc{iForest} can be retrained for every point of the heatmap, so that the explained example is part of the training data. This is the \emph{in-bag} (IB) setting.
Second, if the \textsc{iForest} is not retrained on the example, the setting is \textit{out-of-bag} (OOB).
Third, the training data can contain anomalies, this setting is called \textit{with anomalies} (A), or fourth, it can consist of solely normal points, this setting is called \textit{no anomalies} (No-A).
The last two settings evaluate how the contribution of $X1$ changes when the decision boundary of the \textsc{iForest} also changes.

Figure~\ref{fig:heatmaps} clearly illustrates that our method produces explanations that reflect the real contribution of each attribute, while local-\textsc{Diffi} tends to produce explanations that favor an even contribution of both attributes. 
%This is visible in all plots, as \textsc{Diffi} outputs contributions that are close to 0.5 in most configurations, while our method outputs explanations that show a strong contribution of one feature when a point is mostly anomalous with regard to that feature.
For example, looking at the plots in the first row, the points located at the right edge of the heatmap are mostly anomalous with respect to $X1$. Our method outputs a contribution close to 1 for $X1$, while local-\textsc{Diffi} outputs values close to 0.6, indicating that $X1$ is only slightly more important than $X2$ to explain the anomaly score.
%indicating that the first feature is the one that solely makes these points anomalous. On the other hand, local-\textsc{Diffi} outputs values close to 0.6, indicating that the first feature has a slightly higher contribution in the anomaly score. We argue that the behaviour of loal-\textsc{Diffi} is not correct as it does not reflect the actual impact of the first feature in the anomaly score. 
We argue that local-\textsc{Diffi} is able to correctly rank attributes by their importance, but our method is also able to assign correct values to the attributes' contributions.

To compare the impact of training with or without anomalies, we can compare the first and third columns of Figure~\ref{fig:heatmaps} (IB) or the second and fourth columns (OOB).
%Let us now look at the impact of training with or without anomalies. These differences are depicted by looking at the difference between the first and third columns of Figure \ref{fig:heatmaps}, and the second and forth columns of Figure \ref{fig:heatmaps}.
When using anomalies in the training data, our method accurately identifies the contribution of each attribute over the whole data space.
When no anomalies are in the training data, the explanation is correct mostly where the clusters are located. This is expected as without anomalies, the \textsc{iForest} does not describe the space outside of the clusters, so it provides no information about the attributes' contributions.

Finally, when the \textsc{iForest} is not retrained on the example to explain (OOB), our method outputs contributions that are closer to an even split between both attributes. This is expected as the \textsc{iForest} has less information about which attributes are relevant to isolate this particular point. This effect is more visible when the training data contain no anomalies (third columns against forth column in Figure~\ref{fig:heatmaps}) than when the training data contains anomalies (first column against second column of Figure~\ref{fig:heatmaps}). Training on anomalies provides some information about what attribute is relevant to isolate examples in some parts of the data space.

\section{Conclusion and Future Work}
\label{sec:conclusion}

%While numerous anomaly detection algorithms have been developed, almost no methods exist that explain the output predictions of these algorithms. 
This paper presents a method to explain the predictions of the \textsc{iForest}. It employs a weighting scheme that evaluates how much each attribute of an example contributed to its isolation, forming the basis of the explanation vector for that example.
Our method performs on par with the state-of-the-art \textsc{Shap} method, but is an order of magnitude faster. For future work, we plan to extend our method to global model interpretability, and provide more precise explanations by considering attribute values that make a point anomalous.
\bibliographystyle{unsrt}  
\bibliography{references}

\begin{thebibliography}{10}

\bibitem{Garcia-Teodoro2009-sj}
P~Garc{\'\i}a-Teodoro, J~D{\'\i}az-Verdejo, G~Maci{\'a}-Fern{\'a}ndez, and
  E~V{\'a}zquez.
\newblock Anomaly-based network intrusion detection: Techniques, systems and
  challenges.
\newblock {\em Comput. Secur.}, 28(1):18--28, February 2009.

\bibitem{Thiprungsri2011-jz}
Sutapat Thiprungsri and Miklos~A Vasarhelyi.
\newblock Cluster analysis for anomaly detection in accounting data: An audit
  approach.
\newblock {\em International Journal of Digital Accounting Research}, 11, 2011.

\bibitem{Bolton2001-fl}
Richard~J Bolton, David~J Hand, and {Others}.
\newblock Unsupervised profiling methods for fraud detection.
\newblock {\em Credit scoring and credit control VII}, pages 235--255, 2001.

\bibitem{Lin2005-xz}
J~Lin, E~Keogh, {Ada Fu}, and H~Van~Herle.
\newblock Approximations to magic: finding unusual medical time series.
\newblock In {\em 18th {IEEE} Symposium on {Computer-Based} Medical Systems
  ({CBMS'05})}, pages 329--334, June 2005.

\bibitem{campos2016evaluation}
Guilherme~O Campos, Arthur Zimek, J{\"o}rg Sander, Ricardo~JGB Campello,
  Barbora Micenkov{\'a}, Erich Schubert, Ira Assent, and Michael~E Houle.
\newblock On the evaluation of unsupervised outlier detection: measures,
  datasets, and an empirical study.
\newblock {\em Data mining and knowledge discovery}, 30(4):891--927, 2016.

\bibitem{Miller2019-qb}
Tim Miller.
\newblock Explanation in artificial intelligence: Insights from the social
  sciences, 2019.

\bibitem{molnar2019}
Christoph Molnar.
\newblock {\em Interpretable Machine Learning}.
\newblock 2019.
\newblock \url{https://christophm.github.io/interpretable-ml-book/}.

\bibitem{Liu2008-bj}
F~T Liu, K~M Ting, and Z~Zhou.
\newblock Isolation forest.
\newblock In {\em 2008 Eighth {IEEE} International Conference on Data Mining},
  pages 413--422, December 2008.

\bibitem{domingues2018comparative}
R{\'e}mi Domingues, Maurizio Filippone, Pietro Michiardi, and Jihane Zouaoui.
\newblock A comparative evaluation of outlier detection algorithms: Experiments
  and analyses.
\newblock {\em Pattern Recognition}, 74:406--421, 2018.

\bibitem{cormen2009introduction}
Thomas~H Cormen, Charles~E Leiserson, Ronald~L Rivest, and Clifford Stein.
\newblock {\em Introduction to algorithms}.
\newblock MIT press, 2009.

\bibitem{Ribeiro2016-mt}
Marco~Tulio Ribeiro, Sameer Singh, and Carlos Guestrin.
\newblock {Model-Agnostic} interpretability of machine learning.
\newblock June 2016.

\bibitem{Lundberg2017-ae}
Scott~M Lundberg and Su-In Lee.
\newblock A unified approach to interpreting model predictions.
\newblock In {\em Advances in Neural Information Processing Systems 30}, pages
  4765--4774. 2017.

\bibitem{alvarez2018robustness}
David Alvarez-Melis and Tommi~S Jaakkola.
\newblock On the robustness of interpretability methods.
\newblock {\em arXiv preprint arXiv:1806.08049}, 2018.

\bibitem{Carletti2020-hp}
Mattia Carletti, Matteo Terzi, and Gian~Antonio Susto.
\newblock Interpretable anomaly detection with {DIFFI}: Depth-based feature
  importance for the isolation forest.
\newblock July 2020.

\bibitem{Dua:2019}
Dheeru Dua and Casey Graff.
\newblock {UCI} machine learning repository, 2017.

\end{thebibliography}

\end{document}